%% file: acl_latex.tex
\pdfoutput=1
\documentclass[11pt]{article}

\usepackage[final]{acl}

\usepackage{times}
\usepackage{latexsym}
\usepackage{soul}
\usepackage{multirow}
\usepackage{adjustbox}
\usepackage{xcolor}
\usepackage{textcomp}
\usepackage{graphicx} % 插入图片
\usepackage{subcaption} % 支持子图
\usepackage{amsthm}
\theoremstyle{definition}

\usepackage{bm}
\usepackage{amsfonts}
\usepackage{url}
\usepackage{hyperref}
\usepackage{fancyvrb}
\usepackage{enumitem}
\usepackage{tcolorbox}      % Provides colored box environments
\usepackage{listings}       % Allows inclusion of source code listings
\usepackage[T1]{fontenc}
\usepackage{float}
\usepackage[utf8]{inputenc}

\usepackage{microtype}

\usepackage{inconsolata}

\usepackage{graphicx}
\usepackage{enumitem}
\usepackage{amsmath}
\usepackage{xspace}
\newcommand{\bench}{\texttt{TimerBed}\xspace}
\newcommand{\method}{\texttt{VL-Time}\xspace}
\newcommand{\methodname}{\texttt{VL-Time}}

 \title{A Picture is Worth A Thousand Numbers:\\ Enabling LLMs Reason about Time Series via Visualization}
\author{Haoxin Liu\textsuperscript{\textdagger}\thanks{Correspondence to Haoxin Liu and B. Aditya Prakash}, Chenghao Liu\textsuperscript{\S},  B. Aditya Prakash\textsuperscript{\textdagger}\footnotemark[1] \\ \textsuperscript{\textdagger}Georgia Institute of Technology \quad \textsuperscript{\S}Salesforce Research Asia\\ \textsuperscript{\textdagger}\text{\{hliu763, badityap\}@gatech.edu}, \textsuperscript{\S}\text{chenghao.liu@salesforce.com}
}

\begin{document}
\maketitle
\begin{abstract}

Large language models (LLMs), with demonstrated reasoning abilities across multiple domains, have been largely underexplored for time-series reasoning (TsR), which is ubiquitous in the real world. In this work, we propose \textbf{\bench}, the first comprehensive testbed for evaluating LLMs' TsR performance. Specifically, \bench includes stratified reasoning patterns with real-world tasks, diverse combinations of LLMs and reasoning strategies, and various supervised models as comparison anchors. We perform extensive experiments with \bench, test multiple current beliefs, and observe the initial failures of LLMs in TsR, as evidenced by the ineffectiveness of zero shot (ZST) and performance degradation of few shot in-context learning (ICL). Further, we identify one possible root cause: the numerical modeling of data. To address this, we propose a prompt-based solution \textbf{\method}, with visualization-modeled data and language-guided reasoning. Experimental results demonstrate that \method enables multimodal LLMs to be non-trivial ZST and powerful ICL reasoners for time series, achieving about 140\% average  performance improvement and 99\% average token costs reduction. \textbf{\bench} and \textbf{\method} are available at \url{https://github.com/AdityaLab/DeepTime/}.

\end{abstract}
\input{1_Introduction}

%\input{1-1_realted} % commented out by arXiv, as the file contained no content: 
% $ cat 1-1_realted.tex 
% %\section{Related Works}\label{sec:related}
% this file was removed by the submitter, but an error prevented the update of this main file.
\input{2_Benchmarking}

\input{3_Solution}

\input{4_Experiments}

\section*{Acknowledgements}
This paper was supported in part by the NSF (Expeditions CCF-1918770, CAREER IIS-2028586, Medium IIS-1955883, Medium IIS-2106961, Medium IIS-2403240, PIPP CCF-2200269), CDC MInD program, Meta faculty gift, and funds/computing resources from Georgia Tech and GTRI. This work used the Delta GPU Supercomputer at NCSA of UIUC through allocation CIS240288 from the Advanced Cyberinfrastructure Coordination Ecosystem: Services \& Support (ACCESS) program.
\bibliography{acl_latex}

\include{Z_Appendix}

\end{document}

%% file: 1_Introduction.tex
\section{Introduction}\label{sec:intro}

With the rapid advancement of large language models (LLMs), they have exhibited powerful reasoning abilities with unique interpretability across multiple domains, including logic, mathematics, and symbolic reasoning~\cite{wei2022chain,kojima2022large,achiam2023gpt,dubey2024llama,yang2024qwen2}. Time series, which records the dynamic evolution of data over time, is common in the real world. Time-series reasoning (TsR)~\cite{merrill2024language} has extensive real-world applications, such as pathological diagnosis~\cite{dingwell2000nonlinear}, marine biological monitoring~\cite{baumgartner2011generalized}, and human activity recognition~\cite{yang2015deep}. Therefore, in this paper, we aim to explore the question: \textbf{"Can LLMs be reused for time-series reasoning? If not, how can we enable them?"}

However, this question has remained largely unexplored. The \textbf{first key obstacle is "How to effectively evaluate LLMs on TsR?"}. Existing works~\citep{merrill2024language} face limitations at each component of evaluation: (1) Task Structure and Datasets. They directly mix multiple tasks without a clear taxonomy of reasoning patterns, making it difficult to understand LLMs' limitations in TsR; They rely on synthetic data, which fails to accurately represent real-world scenarios. (2) LLMs Reasoning Strategies: They only adopt the zero-shot setting, ignoring chain-of-thought~\cite{wei2022chain} and few-shot in-context-learning~\cite{brown2020language} strategies, which are widely used to enhance LLMs' reasoning abilities. (3) Comparison Anchors\footnote{In this context, an "anchor" refers to a fixed reference used for performance comparison, which allows researchers to quantify how well LLMs perform.}: They either lack comparison anchor, thus failing to quantify success, or adopt human-level performance, which is costly and difficult to scale to other datasets.

To address these issues, we propose \textbf{\bench}, the first comprehensive and hierarchical test\textbf{\ul{Bed}} for evaluating LLMs \textbf{\ul{r}}easoning capabilities on \textbf{\ul{Time}}-series tasks. Using \bench, we conduct extensive experiments to assess LLMs' TsR effectiveness, test several current beliefs about LLMs, and verify the initial failure of LLMs for TsR.

Inspired by the fact that humans rely on visualization to analyze complex data~\citep{shneiderman2003eyes}, we propose a possible reason for LLMs' failure: the numerical modeling of time-series data, which are typically high-dimensional. We empirically validate this hypothesis and introduce a simple yet effective solution, \textbf{\method}, which leverages \textbf{\ul{V}}isualization for data modeling and \textbf{\ul{L}}anguage for reasoning guidance, thereby unlocking the potential of multimodal LLMs for \textbf{\ul{Time}}-series reasoning. Main contributions are summarized as :
\begin{itemize}
\item We propose \bench, the first comprehensive testbed of evaluating LLMs for TsR. \bench introduces stratified reasoning patterns: simple deterministic reasoning, complex deterministic reasoning, and probabilistic reasoning, with curated real-world tasks. \bench includes diverse combinations of LLMs and reasoning strategies, and various supervised models as comparison anchors.
\item We conduct extensive evaluations with \bench to provide insights and identify a key underlying cause of LLMs' initial failures in TsR: the direct numerical modeling of data. Specifically, numerical modeling results in difficulties in feature extraction and excessively long contexts, which in turn lead to ineffectiveness of zero-shot (ZST) reasoning and performance degradation with in-context learning (ICL), respectively.
\item We propose a prompt-based solution, named \method, to address this limitation. \method employs two steps using LLMs: (1) acts as a domain expert to plan visualization methods and reasoning guidance. (2) performs TsR using visualization-modeled data and language-guided reasoning. \method effectively compresses time-series data, to empower feature extraction and ICL; mimics the human multi-step decision-making process to unlock LLMs' reasoning capabilities.
\item Experimental results show that \method enables multimodal LLMs as non-trivial zero-shot and powerful few-shot reasoners for time series. Specifically, \method with a few demonstration examples can consistently outperform all supervised methods in deterministic reasoning tasks and parts of them in probabilistic reasoning tasks. Compared to direct numerical modeling, \method achieves an average performance improvement of 140\%, with gains up to 433\%, and average token cost reduction of 99\%.
\end{itemize}
In this work, we validate that visualization and few-shot ICL are key to enabling LLMs to reason about time series. We envision visual encoding and ICL can be important ingredients for the next-generation models to move beyond forecasting. Additional related works are provided in Appendix~\ref{sec:related}. Limitations are discussed in Appendix~\ref{sec:limitaion}.

%% file: 2_Benchmarking.tex
\section{\bench: Testbed for LLMs Reasoning about Time Series}\label{sec:testbed}
\begin{figure*}[ht]
     \centering
     \includegraphics[width=0.93\linewidth]{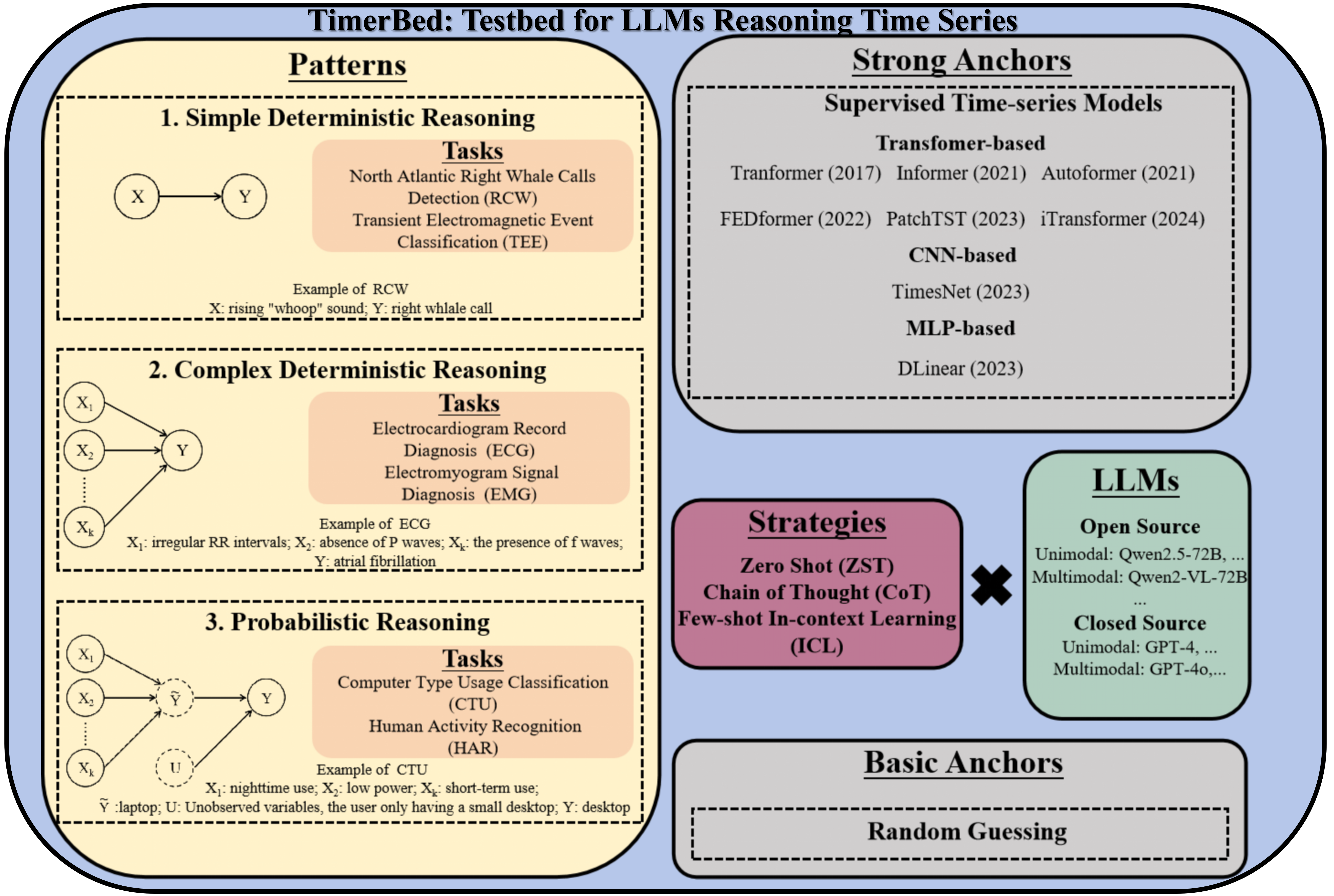}
     \vspace{-0.5mm}
     \caption{Overview of our proposed testbed, \bench, for evaluating LLMs reasoning about time series. \bench defines three patterns of TsR tasks with increasing difficulty: simple deterministic reasoning, complex deterministic reasoning, and probabilistic reasoning. For each reasoning pattern,  \bench matches two real-world tasks and presents an example in the figure. \bench covers four types of LLMs with the corresponding most advanced models and three reasoning strategies for comprehensive evaluation. \bench adopts eight supervised time-series models and random guessing as anchors to quantify the success of LLMs for TsR.}
     \label{fig:suite_overview}
\vspace{-0.5mm}
\end{figure*}
In this section, we first define time-series reasoning following existing LLMs reasoning definitions~\cite{wang2024openr} and then introduce the constructed testbed.

    Let the input numerical time series be $\bm{X}\in\mathbb{R}^{l\times d}$, where $l$ is the series length and $d$ is the feature dimension. Let the textual question be $ \mathbf{Q} $ and the correct answer be $ \mathbf{A} $. The goal of time-series reasoning is to derive a reasoning path $ \bm{P} $, a sequence of intermediate steps, that leads to $\bm{A}$ given $\bm{Q}$ and $\bm{X}$, using zero or a few samples.

As discussed in Section~\ref{sec:intro}, existing evaluation efforts have limitations at every component of benchmarking. To address the question of \textbf{\textit{"How to evaluate and understand the actual TsR capabilities of LLMs?"}}, we present \textbf{\bench}, the first-of-its-kind testbed for evaluating LLM reasoning about time series, as shown in Figure~\ref{fig:suite_overview}. 
We further detail each component of \bench, including reasoning patterns with tasks, LLMs and reasoning strategies, as well as comparison anchors.
\subsection{Stratified Reasoning Patterns with Curated Real-world Tasks}\label{sec:prompt_class}
Inspired by the classic reasoning problem taxonomy~\citep{jaeger1994probabilistic, dechter2022reasoning}, we stratify TsR tasks into three patterns of increasing difficulty: simple deterministic reasoning, complex deterministic reasoning, and probabilistic reasoning. Specifically, "Simple" and "Complex" are defined based on whether multiple factors need to be integrated for reasoning. "Deterministic" and "Probabilistic" are based on whether the relationship between input and output is deterministic. We practically distinguish them based on data annotation methods, i.e., whether labeled by human experts or directly derived from observations. The former establishes a reasoning path from the input data to the label, whereas the latter involves unobserved variables and may introduce spurious correlations. To the best of our knowledge, this is the first stratified evaluation framework for LLMs' TsR capability, supporting a progressive analysis. 

We further mapped two real-world tasks to each pattern for an effective evaluation. Following prior works on benchmarking LLMs' reasoning abilities~\citep{kojima2022large,wei2022chain,sprague2024cot}, we standardized the dataset format as classification tasks. These datasets cover multiple domains (biology, healthcare, electricity, physics, and nature), various task types (binary and multi-class classification, fine-grained and coarse-grained classification), and multiple data types (univariate and multivariate) with varying sizes, precision levels, value ranges, and sequence lengths, thereby enabling a comprehensive evaluation.

Moreover, we curate relevant textual descriptions, including explanations of tasks, data, and labels, which are crucial for LLMs' reasoning. Details of each reasoning pattern and its corresponding tasks are provided below:
\paragraph{Simple Deterministic Reasoning.}\label{sec:RCW}
This pattern is conceptually simple due to the one-to-one correspondence between patterns and their respective classes. We introduce two real-world tasks: north atlantic Right Whale Calls detection (RCW) and Transient Electromagnetic Event classification (TEE). For both tasks, each class is clearly defined by a single feature or pattern, and the labels are annotated by human experts based on input time series, guaranteeing simple deterministic reasoning. The details are provided as follows:

The RCW task \footnote{\url{https://www.kaggle.com/competitions/whale-detection-challenge/data}} contains audio recordings used to identify right whale calls amidst oceanic environmental noise. In this task, a short, rising "whoop" sound serves as both a necessary and sufficient condition for identifying a right whale call. 

The TEE task \footnote{\url{https://www.timeseriesclassification.com/description.php?Dataset=Lightning7}} is sourced from the FORTE satellite, which employs optical and radio frequency instruments to detect transient electromagnetic events associated with lightning. Each event type corresponds to a specific physical phenomenon. For example, the "CG Positive" event type is characterized by a positive charge discharge from cloud to ground, manifested in the power density time series as a distinct radiation spike followed by several hundred microseconds of noise. 
\paragraph{Complex Deterministic Reasoning.}
This pattern is more complex due to the many-to-one mapping between patterns and classes. We introduce two real-world tasks: Electrocardiogram (ECG) Record Diagnosis and Electromyogram (EMG) Signal Diagnosis. These tasks originate from medical scenarios, where the labels are annotated by doctors by integrating multiple patterns from time-series signals, thus ensures both complexity and determinism. The details are as follows:

The ECG task contains single-lead ECG recordings that measure four different cardiac arrhythmias: normal sinus rhythm, atrial fibrillation, other alternative rhythms, and unclassifiable conditions due to excessive noise\footnote{\url{https://physionet.org/content/challenge-2017/1.0.0/}}. Each category requires the integration of multiple factors. For example, doctors diagnose atrial fibrillation by synthesizing multiple typical features, such as irregularly irregular rhythm, absence of P waves, absence of an isoelectric baseline, QRS complexes usually shorter than 120 ms, and the presence of fibrillatory waves~\cite{lip2007management,fuster2006acc}.

The EMG task\footnote{\url{https://physionet.org/content/emgdb/1.0.0/}} comprises muscle responses recordings to neural stimulation, for diagnosing muscular dystrophies and neuropathies. Specifically, it includes single-channel EMG recordings classified into three types: healthy, neuropathy, and myopathy. This task requires a holistic evaluation of multiple patterns. For instance, doctors diagnose neuropathy by considering: fibrillations and positive sharp waves, high-amplitude and long-duration motor unit action potentials, and polyphasic waveforms~\citep{10.1093/med/9780199738687.003.0014,aminoff2004electrophysiologic}.
\paragraph{Probabilistic Reasoning}\label{sec:HAR}
Probabilistic reasoning is challenging due to the uncertain relationship between features and classes, but is common in real-world applications ~\citep{liu2021towards,zhang2022towards}. We introduce two real-world tasks: Human Activity Recognition (HAR) and Computer Type Usage classification (CTU). Probabilistic reasoning is characterized by: (1) labels that are automatically collected, i.e., from observations, and (2) data collected from multiple users, whose heterogeneous behavioral habits significantly impact the results but remain unobserved. Details are as follows:

The HAR task\footnote{\url{https://archive.ics.uci.edu/dataset/240/human+activity+recognition+using+smartphones}} uses wearable sensors on a smartphone to record six daily activities: walking, walking upstairs, walking downstairs, sitting, standing, and lying down. We utilize three sensor channels, which capture body linear accelerations. This dataset was collected from 30 users, whose user-specific behavioral patterns play a crucial role in reasoning but remain unobserved.

The CTU task\footnote{\url{https://www.timeseriesclassification.com/description.php?Dataset=Computers}} aims to infer computer types (desktop or laptop) based on consumer behavior in 24-hour electricity usage. This dataset was collected from 251 users, whose behavioral traits are similarly crucial for reasoning but remain unobserved. More details are in Appendix~\ref{sec:detailofdatasets}.
\subsection{Comprehensive Combination of LLMs and Reasoning Strategies}
Considering the rapid development of LLMs, \bench conducts a comprehensive evaluation by systematically combining various LLMs and reasoning strategies. Specifically, \bench covers four types of LLMs, including open-source unimodal models with the corresponding LLMs: Qwen2.5-72B\footnote{\url{https://qwenlm.github.io/blog/qwen2.5/}} for open-source unimodal LLMs, Qwen2-VL-72B\footnote{\url{https://github.com/QwenLM/Qwen-VL}} for open-source multimodal LLMs, GPT-4\footnote{\url{https://openai.com/index/gpt-4/}} for closed-source unimodal models, and GPT-4o\footnote{\url{https://platform.openai.com/docs/models}} for closed-source multimodal LLMs. The distinction between open-source and closed-source models influences their development approaches, while the choice between unimodal and multimodal architectures affects how they adapt to the underexplored time-series modality.We include multimodal LLMs in our evaluation because they incorporate audio data during pretraining, which shares similarities with time-series data in its sequential nature. We evaluate three widely used reasoning strategies: zero shot (ZST)~\cite{kojima2022large}, zero-shot chain of thought (CoT)~\cite{wei2022chain}, and few-shot in-context learning (ICL)~\cite{brown2020language}.
 
\subsection{Anchors for Quantifying Success}
Considering the differences in reasoning patterns and task difficulty, it is necessary to quantify the success of LLMs using anchors. We include random guessing and eight supervised time-series models as basic and strong anchors, respectively. Specifically, the supervised time-series models cover both time and frequency domains with three main architectures: transformer-based models of time domain~\citep{vaswani2017attention,zhou2021informer,wu2021autoformer,nie2023a,liu2024itransformer} and frequency domain~\citep{zhou2022fedformer}, convolution neural network-based models~\citep{wu2023timesnet}, and multi-layer perceptron-based models~\citep{zeng2023transformers}. More details are provided in Appendix~\ref{sec:supervised}.

\textbf{\bench} and \textbf{\method} are publicly available\footnote{\url{https://github.com/AdityaLab/DeepTime/}}.

%% file: 3_Solution.tex
\section{Evaluations and Insights}\label{conc1}
\subsection{Evaluation Setups}\label{sec:setup}

We represent time-series data as numerical sequences to feed into LLMs, following existing works on prompting LLMs for time-series analysis~\cite{xue2023promptcast,gruver2024large,liu20234lstprompt,dong2024can}. We apply structured formatting instructions, which include mandatory descriptions of tasks, data, and available choices, as well as optional instructions, such as the CoT instruction: "Please solve this problem step by step." For ICL strategies that require context, we randomly sample examples from the original training set to construct the demo set and select the number of demos per category from 1 to 6, following the existing few-shot setting~\citep{jiang2024many}. Details of the prompts are in Appendix~\ref{sec:detailofeval1}. Following existing LLM evaluation works~\citep{wang2023decodingtrust,sun2024trustllm,guha2024legalbench, niklaus2024flawn}, we employ a low temperature setting to ensure more precise outputs. Accuracy is used as the evaluation metric, where a higher value indicates better performance. We run each setting three times and report the average. More setup details are provided in Appendix~\ref{sec:detailofeval2}
\subsection{Results and Insights}\label{sec:results1}
\begin{figure}[t]
     \centering
     \includegraphics[width=0.82\columnwidth]{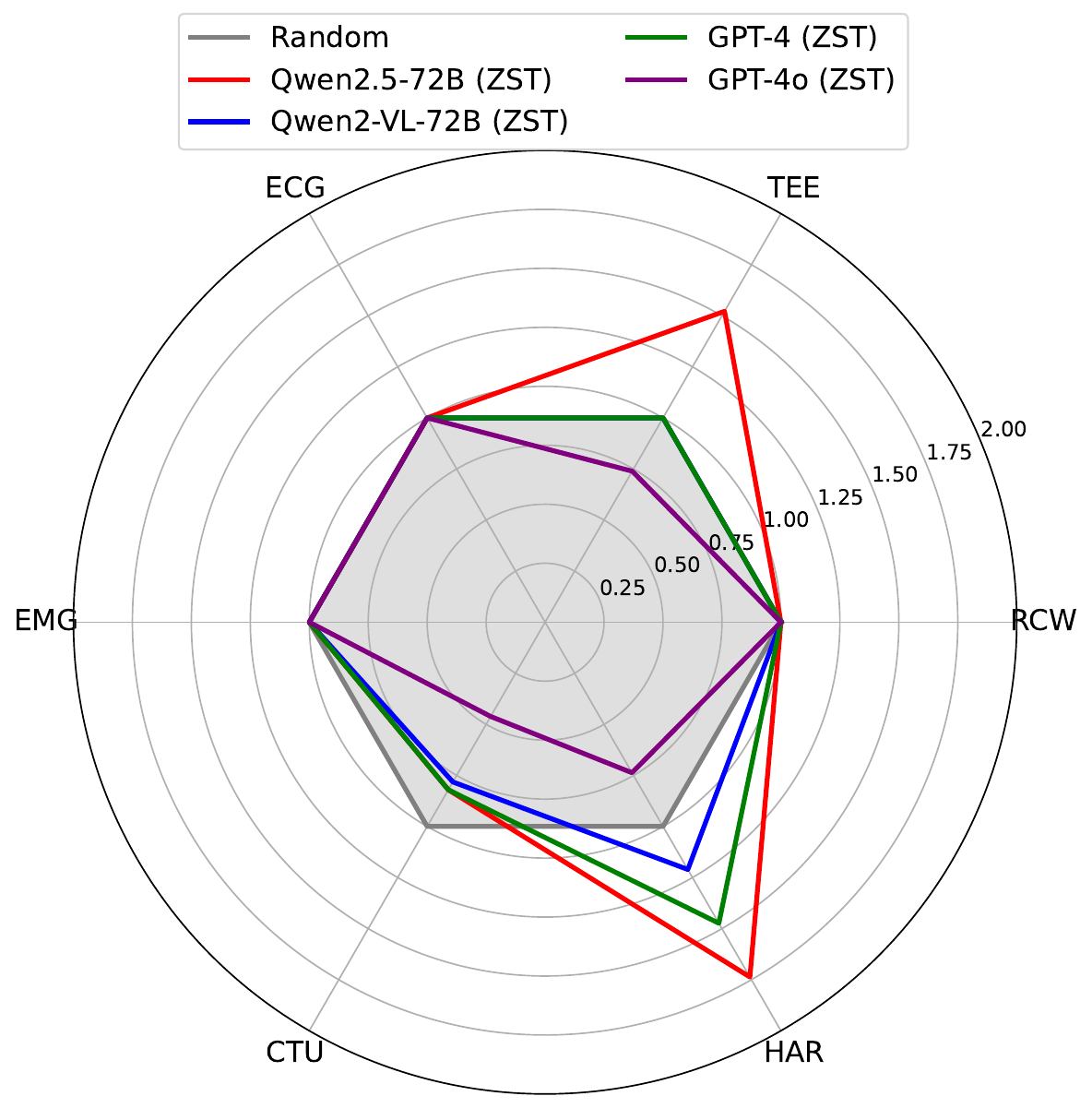}
     \caption{Normalized Results of Zero-Shot Time-series Reasoning. The accuracy is normalized by random guessing. Detailed original results are in Table~\ref{detailofresult1}. LLMs consistently show near-random performance with ZST. }
     \vspace{-0.3mm}
     \label{fig:exp_zst}
\end{figure}

\begin{figure*}[t]
     \centering
     \includegraphics[width=0.8\linewidth]{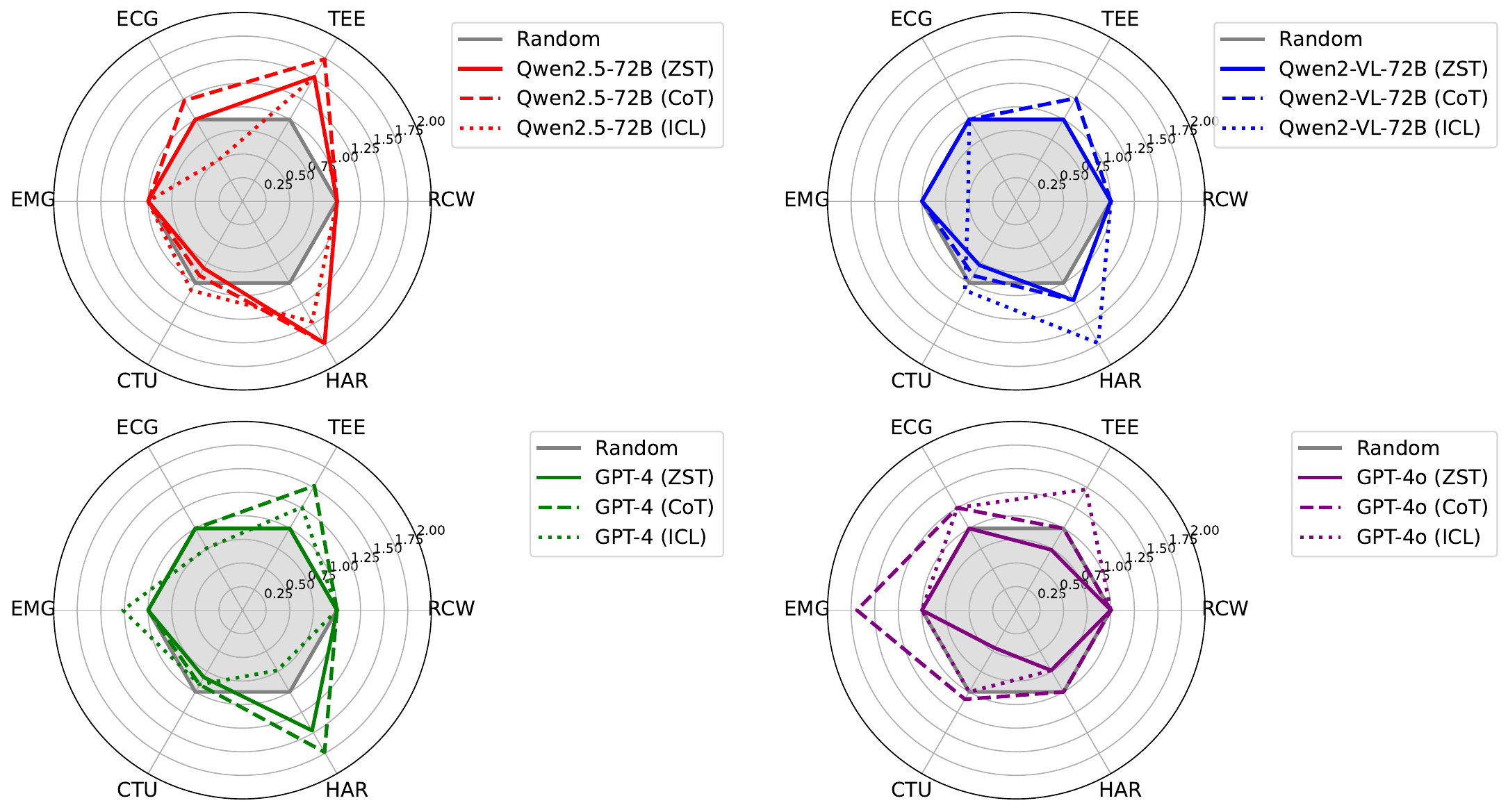}
     \caption{Normalized Results of Chain-of-Thought and Few-shot In-Context-Learning Time-series Reasoning. Each subfigure corresponds to one LLM. The accuracy is normalized by random guessing. Detailed original results are in Table~\ref{detailofresult1}. CoT shows marginal improvement, while ICL leads to performance degradation.}
     \label{fig:exp_fst}
\end{figure*}

We conduct evaluations using \bench with the aforementioned setups to answer the question: \textbf{\textit{"How effective are LLMs for TsR under different reasoning patterns and strategies?"}} We have the following key observations:
\paragraph{Different Reasoning Patterns: ZST Performance is Generally Near-Random.} We visualize ZST results in Figure~\ref{fig:exp_zst}, normalized by the accuracy of random guessing. Detailed original results are in Table~\ref{detailofresult1}. We observe that for all LLMs, their ZST performance across different reasoning patterns is ineffective, often close to or even worse than random guessing. This finding \ul{contradicts the prevailing belief that LLMs are generally non-trivial ZST reasoners}~\cite{kojima2022large}, especially considering that we have provided clear and informative instructions.
\paragraph{Different Reasoning Strategies: CoT > ZST > ICL.} We visualize the results under different reasoning strategies in Figure~\ref{fig:exp_fst}, normalized by the performance of random guessing.  Detailed original results are in Table~\ref{detailofresult1}. 

Across LLMs, we find that CoT consistently outperforms ZST for TsR, validating the rationality of our reasoning tasks, which is \ul{consistent with the prevailing belief that CoT is primarily useful for reasoning tasks}~\citep{sprague2024cot}. 

However, the few-shot ICL strategy for TsR often results in performance degradation, which \ul{contradicts the prevailing belief that LLMs' performance can generally benefit from few-shot ICL}~\citep{brown2020language}.

Additionally, we observe that multimodal LLMs generally perform worse than unimodal LLMs, \ul{consistent with the existing belief about potential degradation of language capabilities in multimodal LLMs}~\citep{lu2024deepseek}.

\section{Failure Analysis: The Numerical Modeling of Data.}\label{sec:aa}
We further investigate \textbf{\textit{"What causes the failures of LLMs for TsR?"}} Focusing on the two aforementioned anomalies, we verify that a key cause is the numerical modeling of time-series data, following most existing works~\cite{xue2023promptcast,gruver2024large,liu20234lstprompt,dong2024can}, detailed as follows:

\paragraph{Numerical Modeled Data $\rightarrow$ Difficulty of Feature Extraction $\rightarrow$ Ineffectiveness of ZST.}

We begin by checking the feature extraction plans of LLMs. Through prompting, we present verbalized feature extraction plans of each LLM for different TsR tasks in Appendix~\ref{sec:pan}. We notice that these plans are similar to each other and, in most cases, are reasonable. This indicates that LLMs possess useful knowledge for zero-shot TsR. However, most of these planned features are difficult to extract from numerical time series and can be classified into four types:

(1) Pictorial Features: For example, "inverted, peaked, or biphasic T-waves" for the ECG task or "paired appearance" for the TEE task are challenging to extract from numerical time series. 

(2) Time-Aware Features: Examples include the "the PR interval, which is normally between 0.12 to 0.20 seconds" for the ECG task and the "noise persisting for several hundred microseconds" for the TEE task. As numerical input of LLMs, time series  naturally lose time stamps, and their lengths, frequently exceeding 1000, further hinder LLMs from accurately tracking time, detailed in Table~\ref{exp_cost2}.

(3) Cross-Dimension Features: For example, "simultaneous changes in X-, Y-, and Z-axis accelerations" for the HAR task. Numerical modeling, which involves concatenating multiple variable series, naturally loses temporal alignment, thus hindering the extraction of cross-dimension patterns. 

(4) Frequency-Domain Features: Such as the "the rising trend in the spectrogram" for the RCW task, visualized in Figure~\ref{fig:RCW1}. Numerical series need transformations like the Fast Fourier Transform~\citep{brigham1988fast} to expose frequency-domain features, which LLMs cannot handle implicitly.

\paragraph{Numerical Modeled Data $\rightarrow$ Excessive Context Length $\rightarrow$ Performance Degradation with ICL.}
The tokenization of LLMs is not optimized for numerical series, leading to an excessive context length. Specifically, one time-series sample typically consists of hundreds of time points, in which each time point is represented by usually more than 12 tokens. The token cost for each dataset is summarized in Table~\ref{exp_cost2}. For example, in the RCW dataset, a single time-series sample requires up to 60K input tokens when processed by GPT-4o. This inefficient tokenization further leads to the performance degradation with ICL due to: 

(1) Challenges in Handling Long Contexts.
With multiple examples included in the context, ICL significantly increases token length and the difficulty of understanding for LLMs~\citep{liu2024lost}.

(2) Limited Capacity for In-Context Examples.
With numerical modeling, the number of time-series examples included in the context is severely limited. For instance, in the RCW dataset, even with an LLM supporting a 128K context length, only one example can be included, despite the dataset having two classes. This leads to biased information and harms performance.

%% file: 4_Experiments.tex
\section{\method: Enabling LLMs Reasoning about Time Series via Visualization}
This section aims to answer: \textbf{\textit{"How can we address the original failure of LLMs in TsR?"}} by focusing on the two observed anomalies: ZST ineffectiveness and ICL performance degradation.
\paragraph{High-Level Idea.} Inspired by the fact that humans rely on visualization to analyze complex data~\citep{shneiderman2003eyes}, we propose to \textbf{replace existing numerical modeling with visualization-based modeling}. In the context of LLMs reasoning about time series, the motivation is detailed as:

(1) Visualization empowers feature extraction. First, visualization makes pictorial features more discernible. Second, visualization can directly reflect time-related features by indicating time stamps using axis indices. Additionally, visualization naturally aligns different variables and uses color and legends to annotate them, thus clearly denoting cross-dimension features. Finally, frequency-domain features can be effectively represented through frequency-domain visualization.

(2) Visualization compresses information, thus reducing token length. Compared to numerical modeling, where token usage is proportional to the data precision and sequence length, visualization serves as an effective compression method. By adjusting the image size, visualization can effectively control token count while largely preserving information. For example, in the RCW task, when processed by GPT-4o, a numerically modeled time-series sample requires up to 60K tokens, whereas only 85 tokens using visualization.
\begin{figure}[t]
     \centering
     \includegraphics[width=\columnwidth]{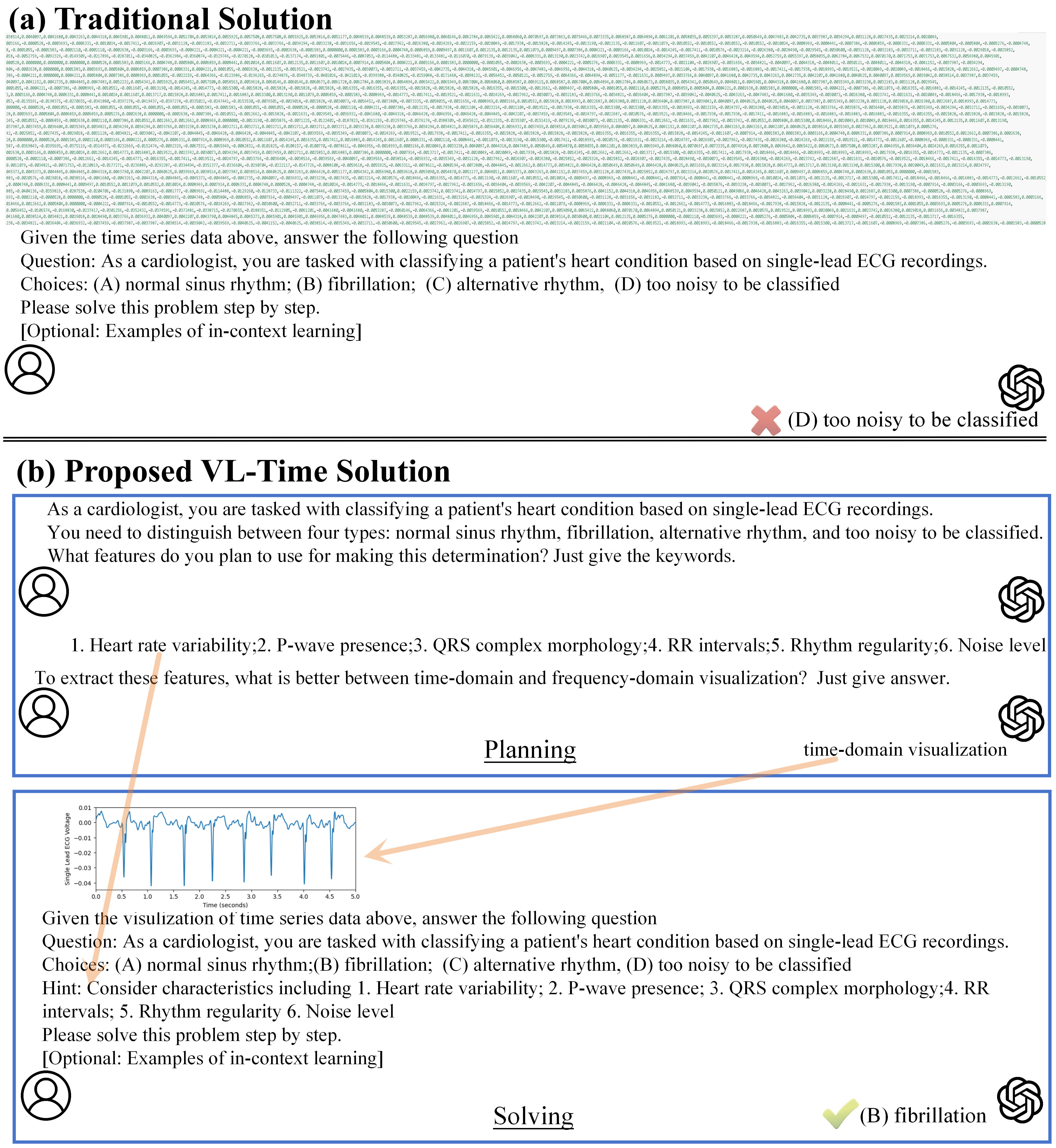}
     \caption{Comparison of existing numerical modeling solution, denoted as "Traditional Solution", and proposed \method. The key difference is that \method replaces numerical modeling with visualization modeling for time-series data, which enhances feature extraction and reduces context length. \method further divides the entire reasoning process into planning and solving stages, mimicking the behavior of human experts. A full example is provided in Section~\ref{sec:full}. For each task, the planning stage needs to be executed only once.}
     \label{fig:method}
     
\end{figure}
\paragraph{Framework and Details.}
As shown in Figure~\ref{fig:method}, we propose a prompt-based solution, named \textbf{\method}, which aims to empower any multimodal LLMs to reason about \textbf{Time} series by data modeling via \textbf{V}isualization and task reasoning guided by \textbf{L}anguage. Specifically, \textbf{\method} adopts a "plan-then-solve" framework~\cite{wang-etal-2023-plan} inspired by human reasoning processes~\cite{wei2022chain} to prompt multimodal LLMs for TsR. 
\begin{table*}[t]
\centering
\resizebox{0.9\linewidth}{!}{
\begin{tabular}{cc|cc|cc|cc} 
\hline
\multicolumn{2}{c|}{Reasoning Pattern}                                                                                                                                                   & \multicolumn{2}{c|}{Simple Deterministic} & \multicolumn{2}{c|}{Complex Deterministic} & \multicolumn{2}{c}{Probabilistic}  \\
\multicolumn{2}{c|}{Reasoning Task}                                                                                                                                                      & RCW             & TEE                     & ECG             & EMG                      & CTU             & HAR              \\
\multicolumn{2}{c|}{Metric}                                                                                                                                                             & ACC(\%)         & ACC(\%)                 & ACC(\%)         & ACC(\%)                  & ACC(\%)         & ACC(\%)          \\ 
\hline
\multicolumn{2}{c|}{Random Guessing}                                                                                                                                                    & 50.00~          & 14.29~                  & 25.00~          & 33.33~                   & 50.00           & 16.67~           \\ 
\hline
\multirow{8}{*}{\begin{tabular}[c]{@{}c@{}}Supervised\\Time-series\\Models\\(8)\end{tabular}} & Transformer                                                                             & 64.12~          & 59.52~                  & 25.00~          & 86.67~                   & 59.20~          & 87.26~           \\
                                                                                              & Autoformer                                                                              & 62.59~          & 26.19~                  & 23.95~          & 46.67~                   & \textbf{67.20~} & 75.04~           \\
                                                                                              & Informer                                                                                & 75.51~          & 59.52~                  & 22.39~          & 66.66~                   & \textbf{67.20~} & 85.83~           \\
                                                                                              & FEDformer                                                                               & 76.59~          & 42.86~                  & 26.40~          & 73.33~                   & 51.60~          & \textbf{89.88~}  \\
                                                                                              & PatchTST                                                                                & 82.11~          & 57.14~                  & 24.82~          & 60.00~                   & 64.00~          & 79.60~           \\
                                                                                              & iTransformer                                                                            & 76.92~          & 21.43~                  & 24.48~          & 46.67~                   & 46.40~          & 89.49~           \\
                                                                                              & TimesNet~                                                                               & 80.23~          & 61.90~                  & 26.20~          & 73.33~                   & 64.00~          & 88.65~           \\
                                                                                              & DLinear                                                                                 & 56.96~          & 47.63~                  & 23.61~          & 46.67~                   & 52.40~          & 48.97~           \\ 
\hline
\multirow{4}{*}{\begin{tabular}[c]{@{}c@{}}Zero-shot~ \\LLMs\end{tabular}}                    & GPT-4o (numeric)~                                                                                 & 50.00~          & 21.43~                  & 25.00~          & 33.33~                   & 45.45~          & 29.17~           \\
                                                                                              & GPT-4o (\method)                                                                          & 70.02~          & 24.88~                  & 26.33~          & 33.33~                   & 50.71~          & 37.50~           \\
                                                                                              & \textbf{\textbf{\textbf{\textbf{\textbf{\textbf{\textbf{\textbf{Improvement}}}}}}}}     & +40.04\%~        & +16.10\%~                & +5.32 \%         & +0.00\%~                  & +11.57\%          & +28.56\%~         \\
                                                                                              & \textbf{\textbf{Win Supervised}}                                                       & ~3/8            & ~1/8                    & ~7/8            & 0/8                      & ~1/8            & 0/8              \\ 
\hline
\multirow{4}{*}{\begin{tabular}[c]{@{}c@{}}Few-shot ICL \\LLMs\end{tabular}}                  & GPT-4o (numeric)                                                                                 & 50.00~          & 35.71~                  & 31.25~          & 33.33~                   & 50.00~          & 12.50~           \\
                                                                                              & GPT-4o (\method)                                                                          & \textbf{91.03~} & \textbf{64.29~}         & \textbf{43.75~} & \textbf{91.67~}          & 63.64\textbf{~} & 66.67~           \\
                                                                                              & \textbf{\textbf{\textbf{\textbf{Improvement}}}}                                         & +82.06\%~        & +80.03\%~                & +40.00\%~        & +175.04\%~                & +27.28\%~        & +433.36\%        \\
                                                                                              & \textbf{\textbf{\textbf{\textbf{\textbf{\textbf{\textbf{\textbf{Win Supervised}}}}}}}} & 8/8             & 8/8                     & 8/8             & 8/8                      & 4/8             & 1/8              \\
\hline
\end{tabular}}
\caption{Performance comparison on \bench: (1) \method enables GPT-4o to perform non-trivial zero-shot TsR, with up to 40\% relative improvement and consistently outperforming random guessing on all tasks. (2) \method enables GPT-4o to achieve powerful few-shot TsR, with up to 433\% relative improvement, surprisingly surpassing all supervised models in simple and complex deterministic TsR, and matching their performances in probabilistic TsR. Here, few-shot refers to fewer than sixexamples per class. We bold the best result for each task.
}
\label{table:compare}
\end{table*}

The Planning Stage. \method prompts LLMs to act as domain experts 
 and make two key plans based on the task description: (1) choosing between time-domain or frequency-domain visualization, and (2) proposing the reasoning clues. To facilitate effective feature extraction, our visualization designs include time stamps on the x-axis, semantic labels on the y-axis, and different colors and textual legends for each variable or frequency. Figure~\ref{fig:HAR} and~\ref{fig:RCW1} show examples of our visualization. Notably, the planning stage only needs to run once for each task, i.e., shared across samples.

The Solving Stage. \method prompts multimodal LLMs to perform TsR using visualized data (i.e., "Picture") and verbalized reasoning clues (i.e., "Hints") derived from the planning stage. 
\section{Experiments for Proposed Solution}\label{conc2}
This section aims to answer the question:\textbf{\textit{"How well can \method unlock multimodal LLMs' reasoning ability for time series?"}}
\subsection{Performance Comparison}
We evaluate \method on \bench, following the same experimental setup as in Section~\ref{sec:results1}. We use GPT-4o as the default representative model, while results for other LLMs are presented in Section~\ref{exp:morellm}. To avoid confusion, we use GPT-4o (numeric) to denote existing numerical modeling solution evaluated in Section~\ref{sec:results1}.
We use the low resolution option of GPT-4o API. For supervised models, we retain each dataset's original train-test split or used an 8:2 split of the original training set when the test set lacked available labels. As shown in Table~\ref{table:compare}, we make the following observations:

\paragraph{\method enables multimodal LLMs as non-trivial zero-shot time-series reasoners.}
First, \method significantly improves GPT-4o's performance by an average of 17\% and up to 40\%. Additionally, \method allows GPT-4o to surpass random guessing on all tasks and outperform part of supervised models on 4/6 tasks. Compared to GPT-4o (numeric)'s random-level performance, \method enables a non-trivial improvement.
\paragraph{\method enables multimodal LLMs as powerful few-shot in-context time-series reasoners.}
First, \method boosts GPT-4o's performance by an impressive average of 140\% and up to 433\%. Besides, \method enables GPT-4o to outperform all supervised time-series models on all tasks with simple and complex deterministic reasoning patterns. Moreover, \method even matches the performance of supervised models (surpassing parts of them) on all tasks with probabilistic reasoning patterns. Recall that probabilistic reasoning demands more data to capture uncertainty. However, \method remarkably enables multimodal LLMs to achieve strong performance with fewer than six examples per class, demonstrating unique data efficiency.
\paragraph{Cost Comparison.}
\method introduces no additional training cost. The cost difference mainly comes from the data modeling approach: visual modeling in \method and numerical modeling in traditional prompts. Thus, we report the input token cost and per-sample token count in Table~\ref{exp_cost1} and \ref{exp_cost2} with GPT-4o. Results show that \method significantly reduces token count and costs, requiring only \textbf{about 1\% of numerical modeling's cost}.
\subsection{Ablation Studies}
We choose the zero-shot RCW\footnote{The Right Whale Calls detection task in Section~\ref{sec:RCW} } and HAR\footnote{The Human
Activity Recognition task in Section~\ref{sec:HAR} } tasks. 
\paragraph{Effectiveness of the Planning Stage.} To validate the effectiveness of the first stage, i.e., planning, we use only the second stage, i.e., planning, referred to as \methodname$\backslash$\texttt{Planning}. Due to the lack of a decision between time- and frequency-domain visualization, \methodname$\backslash$\texttt{Planning} uniformly adopts the more common time-domain visualization.

\begin{figure}[!ht]
     \centering
     \includegraphics[width=\columnwidth]{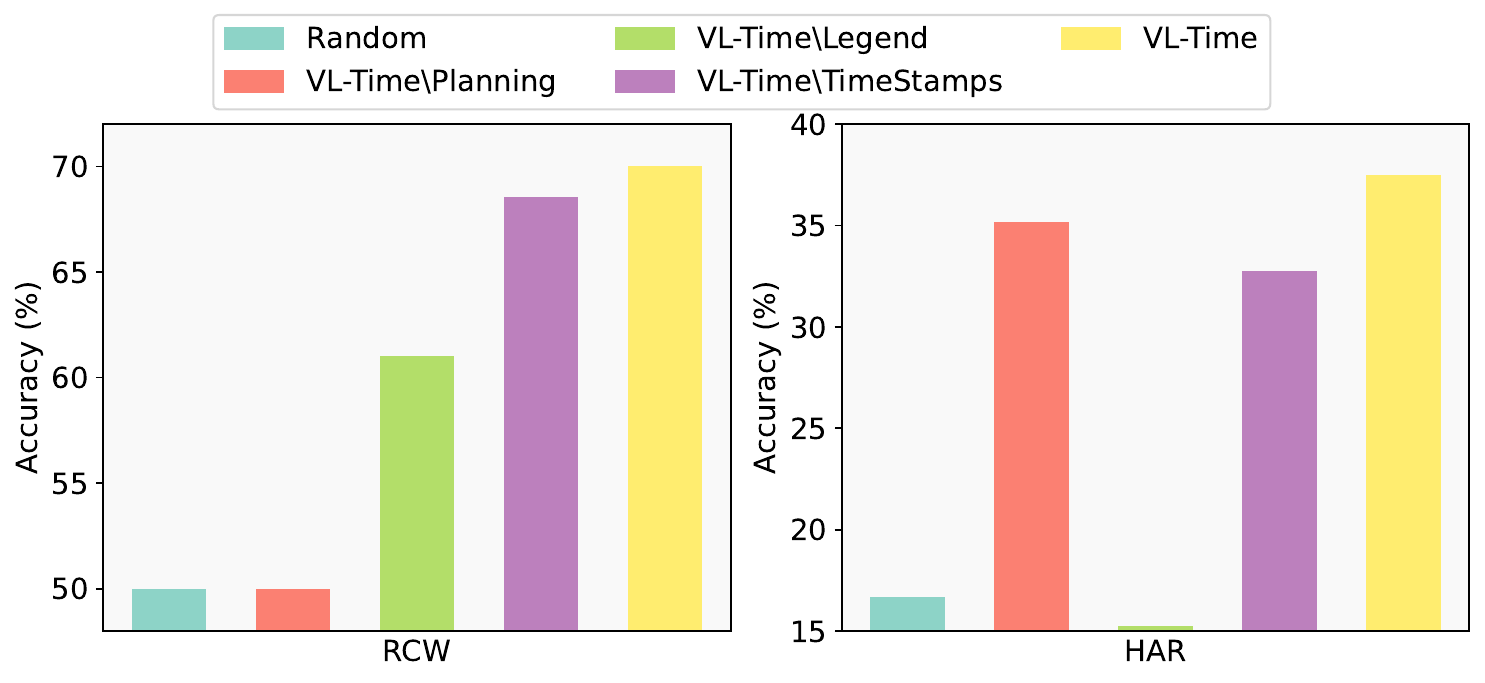}
     \caption{Ablation Study of \method. The planning stage and visualization designs, including the textual legend and timestamps, are all validated as effective.}
     \label{fig:exp_Abl}
\vspace{-0.1mm}
\end{figure}

As shown in Figure~\ref{fig:exp_Abl}, the planning stage proves effective. Firstly, \methodname$\backslash$\texttt{Planning} performs close to random on RCW, highlighting the importance of planning for time- or frequency-domain visualizations, especially since RCW relies on frequency features. Moreover, for HAR, a time-domain task, \methodname$\backslash$\texttt{Planning} also shows performance degradation, demonstrating that feature hints from the planning stage play a crucial role in guiding the solving stage.
\paragraph{Effectiveness of the Visualization Design.} We focus on two key visualization components in \method: the textual legend and annotated timestamps, which correspond to two ablation versions, \methodname$\backslash$\texttt{Legend} and \methodname$\backslash$\texttt{TimeStamps}, respectively. Specifically, RCW uses legends to map colors to decibel levels, and HAR uses legends to map curves to the x, y, and z axes of acceleration. The visualization demos for RCW and HAR are provided in Figure~\ref{fig:RCW1} and~\ref{fig:HAR}. 

As shown in Figure~\ref{fig:exp_Abl}, these two visualization designs demonstrate their effectiveness.: (1) \methodname$\backslash$\texttt{Legend} shows a significant performance drop, particularly for HAR, where identifying the x, y, and z axes is crucial for activity recognition. (2) \methodname$\backslash$\texttt{TimeStamps} shows consistent performance degradation, indicating that capturing time-aware features is helpful.
\begin{figure}[!ht]
    \centering
    \begin{subfigure}[b]{\linewidth}
        \includegraphics[width=\linewidth]{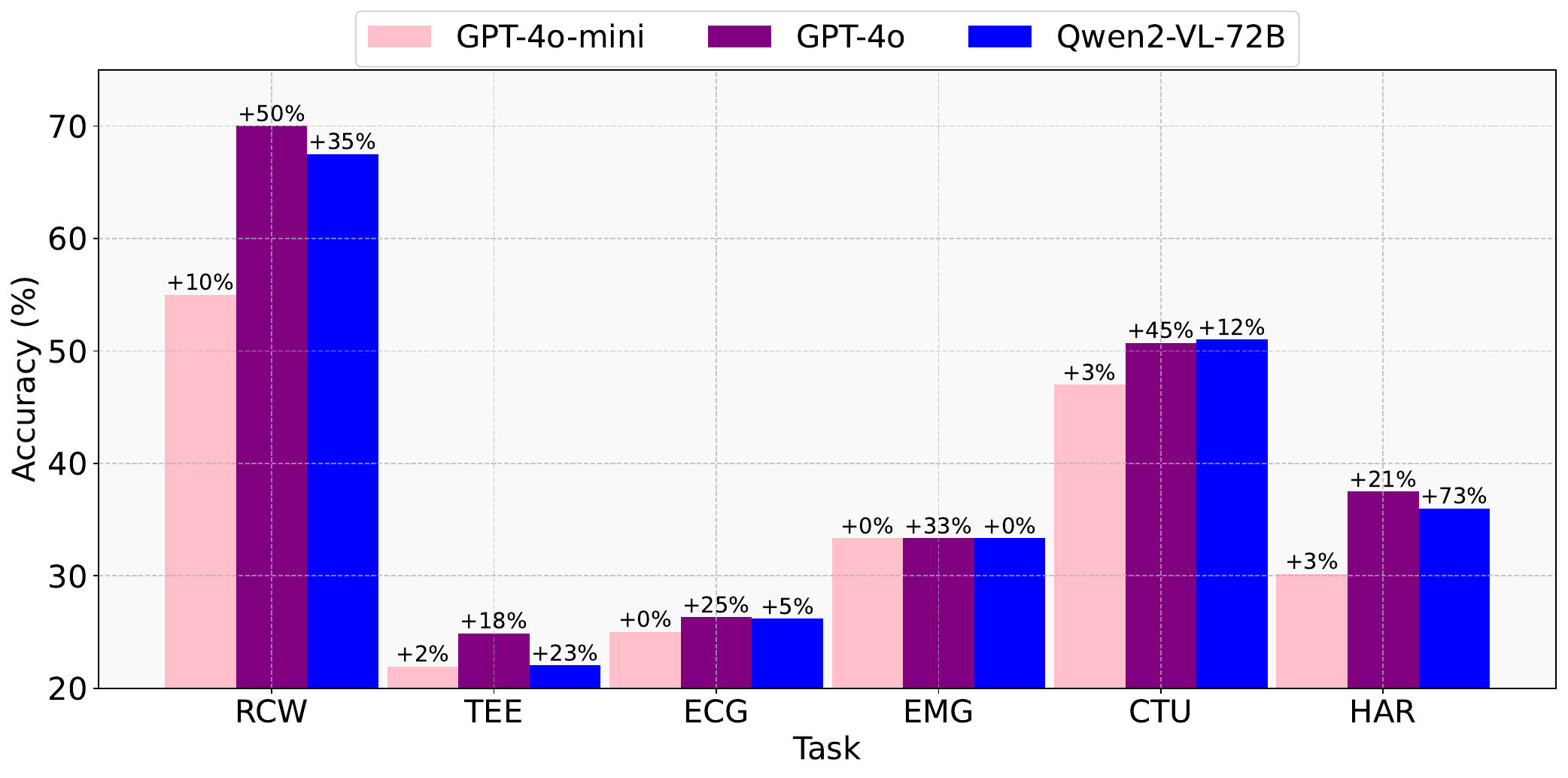}
        \caption{Results of LLMs selection with ZST.}
        \label{fig:exp_LLMs_zero}
    \end{subfigure}
    \begin{subfigure}[ht]{\linewidth}
        \includegraphics[width=\linewidth]{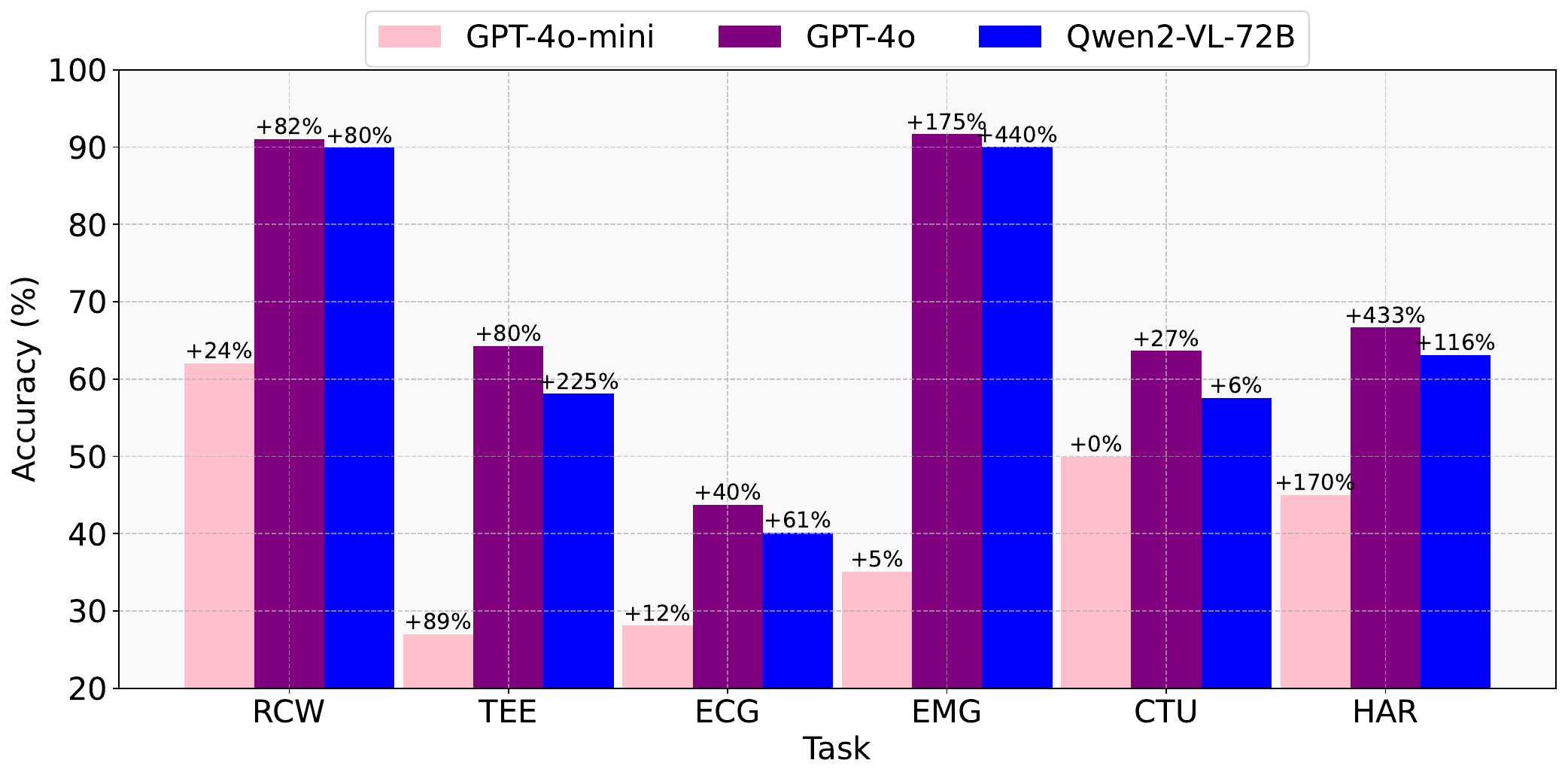}
        \caption{Results of LLMs selection with ICL.}
        \label{fig:exp_LLMs_few}
    \end{subfigure}
    \caption{Results of LLMs selections. We label the improvements of \method\ over the numerical modeling at the top of bars. We observe the consistent effectiveness of \method, the importance of model scale, and the gap between open- and closed-source models.}
    \label{fig:exp_LLMs_combined}
\end{figure}
\subsection{Exploratory Studies}
\paragraph{Multimodal LLMs Selection within \method.}\label{exp:morellm}
We analyze the impact of LLM selection to answer three questions: (1) Is \method effective across different LLMs? (2) How does model size affect TsR performance? (3) What is the gap between open-source and closed-source models? 

To answer these questions, we introduce two additional multimodal LLMs: GPT-4o-mini for Question (2) and Qwen2-VL-72B for Question (3). 

As shown in Figure~\ref{fig:exp_LLMs_zero} and \ref{fig:exp_LLMs_few}, we make the following observations: (1) \method is effective across different multimodal LLMs.
(2) LLM's TsR capability relies on model size. (3) State-of-the-art open-source multimodal LLMs still lag behind closed-source models on TsR tasks.

Results of the open-source Llama-3.2-11B and Llama-3.2-90B are detailed in Section~\ref{exp:morellm}. We have the similar observations.
\section{Conclusion}

In this work, we explore LLMs reasoning about time series with a two-fold contribution: evaluation and methodology. We present the first comprehensive evaluation suite, conduct extensive evaluations, provide insights and verify the initial failures. We identify the direct numerical data modeling as one key issue and propose a prompt-based solution with visualized data modeling. We empirically validate that visualization of data and few-shot ICL are key to enabling LLMs reasoning about time series. We expect that visual encoding and ICL could be important ingredients for future time-series models.

%% file: Z_Appendix.tex
\appendix
\newpage
\section*{Appendix}
\section{Additional Related Works}\label{sec:related}

LLMs have a wide range of applications across various domains~\citep{jin2024mm,xiao2024proteingpt}. Here, we introduce related works on prompting LLMs for time series analysis. These works~\citep{xue2023promptcast,gruver2024large,liu20234lstprompt,dong2024can} mainly model time series numerically as language inputs to LLMs. This approach is somewhat limited by the fact that LLMs at that time are mainly unimodal. With the recent rapid development of multimodal  LLMs~\citep{yin2023survey,zhang2024out} and multimodal time-series analysis~\citep{liu2024timemmd}, our proposed solution aims to demonstrate that visualization modeling is the key to  LLMs reasoning about time series.

Recently, limited works~\citep{merrill2024language,chow2024towards} try to explore the use of LLMs for time-series reasoning, with preliminary conclusions indicating that LLMs still struggle with zero-shot reasoning on time series. However, our work demonstrates that, for current LLMs, visualization of data and few-shot in-context learning are critical for effective reasoning about time series, as verified in Section~\ref{sec:intro}. Furthermore, our work addresses the limitations in evaluation present in existing studies, as discussed in Section~\ref{conc2}.

\section{Limitations}\label{sec:limitaion}
According to the three levels of causality defined by \citeauthor{pearl2018book}, this paper focuses on reasoning problems at the basic "Association" level. For reasoning problems at the further "Intervention" and "Counterfactual" levels, we look forward to future works, especially in addressing the lack of suitable datasets. 

Besides, all reasoning tasks in this work are formalized as classification tasks. For other forms of time series analysis, it remains unclear whether task performance can reveal reasoning ability. Specifically, time-series forecasting~\citep{liu2024time,jin2024timellm,LPTM} and imputation~\citep{du2024tsi} focus on generating specific future numerical sequences, which are typically "open-ended." For example, when forecasting future influenza infection trends, any unforeseen event within the prediction window, such as a disease outbreak, might impact the final future values.~\citep{mathis2024evaluation}. We expect future research to further explore how to reasonably evaluate reasoning ability using these task formats.

Moreover, this work focuses on a prompt-based solution to unlock the ability of multimodal LLMs to reason about time series, which is simple but effective. We expect that our evaluation suite, insights, and solution will inspire future fine-tuning-based approaches and more general solutions for both unimodal and multimodal LLMs.

Additionally, this work leverages the vision-text alignment capability of multimodal LLMs to achieve time series-text alignment. Our concurrent work~\citep{cai2024timeseriesexam} also evaluated the superiority of visual modeling in feature alignment. However, this approach focuses on repurposing existing LLMs rather than serving as the optimal solution. We look forward to future research on time-series-specific multimodal models, including improvements in accuracy and safety,such as addressing hallucination issues\citep{agarwal2024medhalu}.

\section{Detailed Evaluation Results}
Detailed Evaluation Results are provided in Table~\ref{detailofresult1}.

\section{Details of Datasets}\label{sec:detailofdatasets}
Statistics of original dataset are provided in Table~\ref{Tab:datasets}.
\begin{table*}[!ht]
\centering
\begin{tabular}{ccccc} 
\hline
Dataset & Number of Variable & Length of Series & Number of Class  &Number of Samples \\ 
\hline
RCW     & 1                  & 4000             & 2           & 30,000     \\
TEE     & 1                  & 319              & 7           &143     \\
ECG     & 1                  & 1500             & 4           &43,673     \\
EMG     & 1                  & 1500             & 3           &205     \\
CTU     & 1                  & 720              & 2           &500     \\
HAR     & 3                  & 206              & 6          &10,299      \\
\hline
\end{tabular}
\caption{Statistics of Original Datasets.}\label{Tab:datasets}
\end{table*}
\begin{table*}[!ht]
\centering
\begin{tabular}{cc|cccccc} 
\hline
\multicolumn{2}{c|}{Task}              & RCW     & TEE     & ECG     & EMG     & CTU     & HAR      \\ 
\hline
\multicolumn{2}{c|}{Metric}            & ACC(\%) & ACC(\%) & ACC(\%) & ACC(\%) & ACC(\%) & ACC(\%)  \\ 
\hline
\multicolumn{2}{c|}{Random Guessing}   & 50.00~  & 14.29~  & 25.00~  & 33.33~  & 50.00   & 16.67~   \\ 
\hline
\multirow{3}{*}{Qwen2.5-72B}  & ZST        & 50.00~  & 21.43~  & 25.00~  & 33.33~  & 40.91~  & 29.17~   \\
                          & CoT        & 50.00~  & 25.00~  & 30.77~  & 33.33~  & 45.45~  & 29.17~   \\
                          & ICL        & 50.00~  & 21.43~  & 12.50~  & 33.33~  & 54.55~  & 25.00~   \\ 
\hline
\multirow{3}{*}{Qwen2-VL-72B} & ZST        & 50.00~  & 14.29~  & 25.00~  & 33.33~  & 38.89~  & 20.83~   \\
                          & CoT        & 50.00~  & 17.86~  & 25.00~  & 33.33~  & 45.45~  & 20.83~   \\
                          & ICL        & 50.00~  & 17.86~  & 25.00~  & 16.67~  & 54.55~  & 29.17~   \\ 
\hline
\multirow{3}{*}{GPT-4}     & ZST        & 50.00~  & 14.29~  & 25.00~  & 33.33~  & 40.91~  & 25.00~   \\
                          & CoT        & 50.00~  & 21.43~  & 25.00~  & 33.33~  & 45.45~  & 29.17~   \\
                          & ICL        & 50.00~  & 17.86~  & 18.75~  & 41.67~  & 45.45~  & 12.50~   \\ 
\hline
\multirow{3}{*}{GPT-4o}   & ZST        & 50.00~  & 10.71~  & 25.00~  & 33.33~  & 22.73~  & 12.50~   \\
                          & CoT        & 50.00~  & 14.29~  & 31.25~  & 55.56~  & 54.55~  & 16.67~   \\
                          & ICL        & 50.00~  & 35.71~  & 31.25~  & 33.33~  & 50.00~  & 12.50~   \\
\hline
\end{tabular}
\caption{Detailed Results of Time-series Reasoning without Normalization.}\label{detailofresult1}
\end{table*}
\section{Details of Prompts}\label{sec:detailofeval1}
Prompts used for Zero-Shot,  Zero-Shot Chain-of-Thought, Few-shot In-Context Learning Setting are detailed in Figure~\ref{fig:p1}, Figure~\ref{fig:p2} and Figure~\ref{fig:p3} correspondingly.
\begin{figure*}[h]
    \centering
    \begin{tcolorbox}
    \begin{lstlisting}[breaklines=true]
prompt = f"""
{TimeSeries_Number}. 
Given the time series data above, 
answer the following question
using the specified format.
    Question: {task_description}
    Choices: {str(class_description)}
    Please respond with the following format:
    ---BEGIN FORMAT TEMPLATE---
    Answer Choice: [Your Answer Choice Here]
    ---END FORMAT TEMPLATE---
    Do not deviate from the above format. 
    Repeat the format template for the answer.
"""
\end{lstlisting}
    \end{tcolorbox}
    \caption{Prompt used for Zero-Shot Setting.}
    \label{fig:p1}
\end{figure*}
\begin{figure*}[h]
    \centering
    \begin{tcolorbox}
    \begin{lstlisting}
prompt = f"""
{TimeSeries_Number}. 
Given the time series data above, 
answer the following question
using the specified format. 
    Question: {task_description}
    Choices: {str(class_description)}
    Please solve this problem step by step.
    Please respond with the following format:
    ---BEGIN FORMAT TEMPLATE---
    Answer Choice: [Your Answer Choice Here]
    ---END FORMAT TEMPLATE---
    Do not deviate from the above format. 
    Repeat the format template for the answer.
"""
\end{lstlisting}
    \end{tcolorbox}
    \caption{Prompt used for Zero-Shot Chain-of-Thought Setting.}
    \label{fig:p2}
\end{figure*}
\begin{figure*}
    \centering
    \begin{tcolorbox}
    \begin{lstlisting}
prompt = f"""
for demo in demo_examples:
    prompt += f"""{TimeSeries_Number}. 
    Given the time series data above, 
    answer the following question
    using the specified format.
    Question: {task_description}
    Choices: {str(class_description)}
    Answer Choice: {demo.answer}
"""
prompt += f"""
{TimeSeries_Number}. 
Given the time series data above, 
answer the following question
using the specified format. 
    Question: {task_description}
    Choices: {str(class_description)}
    Please solve this problem step by step.
    Please respond with the following format:
    ---BEGIN FORMAT TEMPLATE---
    Answer Choice: [Your Answer Choice Here]
    ---END FORMAT TEMPLATE---
    Do not deviate from the above format. 
    Repeat the format template for the answer.
"""
\end{lstlisting}
    \end{tcolorbox}
    \caption{Prompt used for Few-shot In-Context Learning Setting.}
    \label{fig:p3}
\end{figure*}

\subsection{Details of Task Description}\label{sec:prompt_task}
\subsubsection{RCW}
Play the role of a marine biology expert: is there a right whale call in the record?
\subsubsection{TEE}
Based on the power density time series data and select the transient electromagnetic event that best matches. The FORTE satellite detects transient electromagnetic events associated with lightning using a suite of optical and radio-frequency (RF) instruments. There are 7 event types. CG Positive Initial Return Stroke: A positive charge is lowered from a cloud to the ground. The characteristic feature of this type of event in the power density time series is a sharp turn-on of radiation, followed by a few hundreds of microseconds of noise; IR Negative Initial Return Stroke: A negative charge is lowered from a cloud to ground. The power waveform slowly ramps up to a level known as an attachment point, where a large surge current causes the VHF power to 'spike'. This attachment is followed by an exponentially shaped decline in the waveform.; SR Subsequent Negative Return Stroke: A negative charge is lowered from a cloud to ground. As the name implies, subsequent return strokes come after initial return strokes. Note that subsequent positive return strokes don't exist. I Impulsive Event: Typically an intra-cloud event characterized by a sudden peak in the waveform. I2 Impulsive Event Pair: Another intra-cloud event characterized by sudden peaks in the waveform that come in closely separated pairs. These are also called TIPPs (Trans-Ionospheric Pulse Pairs). KM Gradual Intra-Cloud Stroke: An intra-cloud event which increases in power more gradually than an impulsive event. O Off-record: 800 microseconds was not enough to fully capture the lightning event.

\subsubsection{ECG}
As a cardiologist, you are tasked with classifying a patient's heart condition based on single-lead ECG recordings.

\subsubsection{EMG}
As an Electromyograms (EMG) analysis expert, you are tasked with determining the type of the subject based on the EMG record.

\subsubsection{CTU}
Play as a computer energy consumption analysis expert, please correctly determine whether this computer is a desktop or a laptop based on the 24-hour power consumption data.

\subsubsection{HAR}
As a human activity recognition expert, you are tasked with determining the type of activity performed by the subject based on the accelerometer record series along the x, y, and z axes over time.
\section{Full Sample}\label{sec:full}
Here, we provide a complete input example from an ECG task. Other tasks can be easily implemented by replacing the corresponding task description in Section~\ref{sec:prompt_task} and the class descriptions from Section~\ref{sec:prompt_class}.
\subsection{Input of the planning stage}
As a cardiologist, you are tasked with classifying a patient's heart condition based on single-lead ECG recordings. 
You need to distinguish between four types: normal sinus rhythm, fibrillation, alternative rhythm, and too noisy to be classified. What features do you plan to use for making this determination? Just give the keywords.

To extract these features, what is better between time-domain and frequency-domain visualization?  Just give answer.
\subsection{Input of the solving stage}
<IMAGE> Given the visualization of time series data above, 
answer the following question
Question: As a cardiologist, you are tasked with classifying a patient's heart condition based on single-lead ECG recordings. 
Choices: (A) normal sinus rhythm;
               (B) fibrillation; 
               (C) alternative rhythm,
               (D) too noisy to be classified
Hint: Consider characteristics including 1. Heart rate variability; 2. P-wave presence; 3. QRS complex morphology;4. RR intervals; 5. Rhythm regularity
6. Noise level
Please solve this problem step by step.
[Optional: Examples of in-context learning]
\begin{figure}[!ht]
     \centering
     \includegraphics[width=0.78\columnwidth]{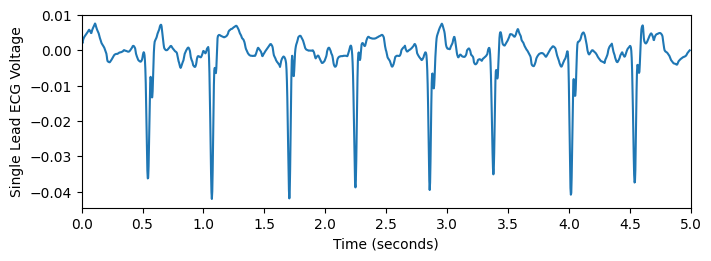}
     \caption{<IMAGE>: The visualization used in this full sample.}
\end{figure}
\section{Additional Experiments}
\begin{figure}[!ht]
    \centering
    \begin{subfigure}[b]{\linewidth}
        \includegraphics[width=\linewidth]{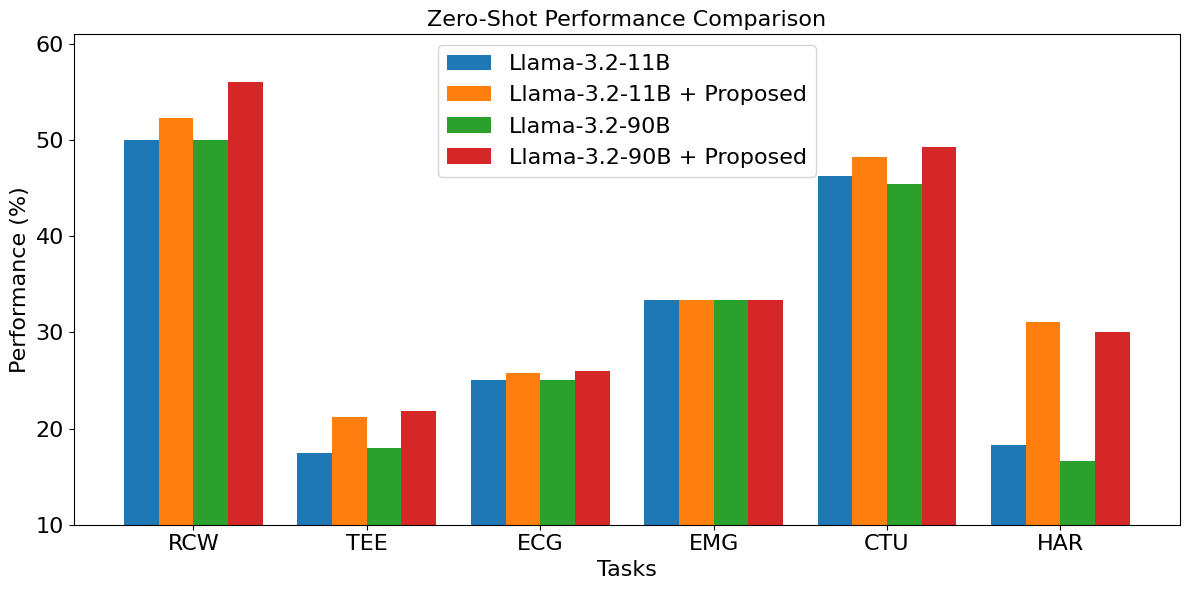}
        \caption{Results of zero shot.}
    \end{subfigure}
    \begin{subfigure}[b]{\linewidth}
        \includegraphics[width=\linewidth]{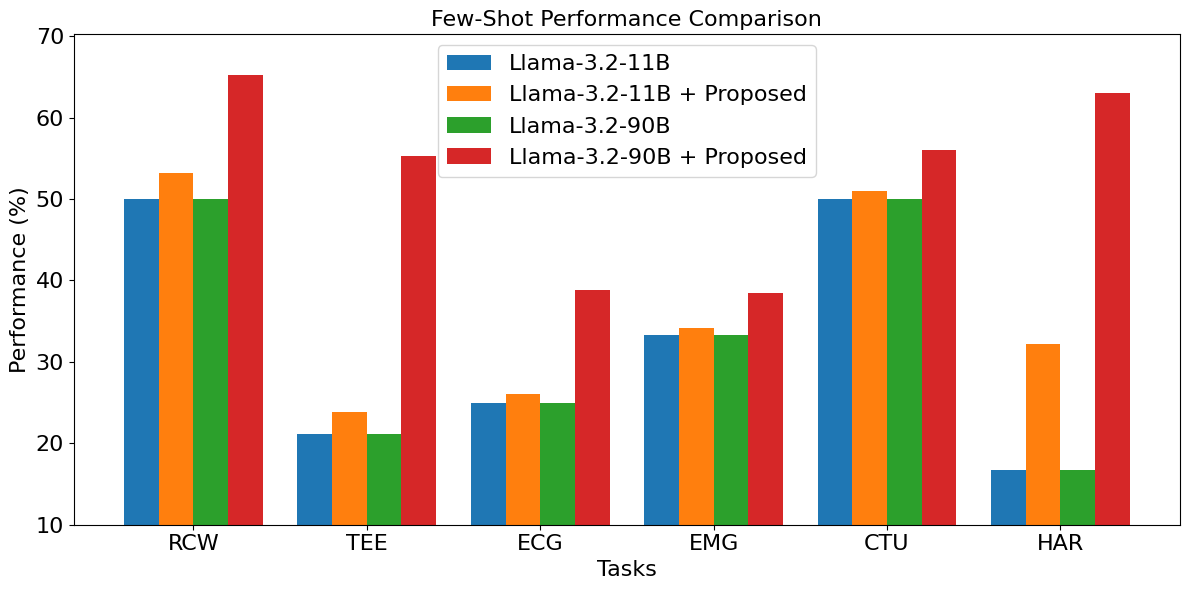}
        \caption{Results of few shot.}
    \end{subfigure}
    \caption{Results on Llama-3.2.}
    \label{fig:Llama}
\end{figure}

Additional experiments on Llama-3.2-11B and Llama-3.2-90B are shown in Figure~\ref{fig:Llama}.
\section{More Details of Setup}\label{sec:detailofeval2}
To prevent models from abstaining (which is common in ICL setting with numerical modeling but rarely for our \method), we rerun the query until an answer is provided. 
We follow the most recent evaluation work on multimodal LLMs ~\citep{jiang2024many} and LLMs for time series~\citep{gruver2024large} to perform sampling, aiming to control class balance and evaluation costs, i.e., evaluating LLMs using a subset of the original test set.

\section{Included Eight Supervised Time-Series Models}\label{sec:supervised}
\begin{itemize}
    \item \textbf{Transformer}\citep{vaswani2017attention}, a classic sequence-to-sequence model based on multi-head attention mechanism.
    \item \textbf{Informer}\citep{zhou2021informer}, an advanced Transformer designed to tackle long-term modeling problem with sparse attention layers and self-attention distilling.
    \item \textbf{Autoformer}\citep{wu2021autoformer}, a Transformer-based model that keeps encoder-decoder structure but alters attention computations by auto-correlation mechanism in order to benefit long-term modeling.
    \item \textbf{FEDformer}\citep{zhou2022fedformer}, a Transformer that explicitly uses Fourier decomposition results to enhance long-term modeling ability.
      \item \textbf{PatchTST}\citep{nie2023a}, a transformer-based model using segmentation of time series and channel-independence design.
    \item \textbf{iTransformer}\citep{liu2024itransformer}, an inverted Transformer that tokenizes multivariate time-series at each timestamp.

    \item \textbf{DLinear}\citep{zeng2023transformers}, a linear model that performs modeling by a direct regression upon historical time series with a one-layer linear model.
    \item \textbf{TimesNet}\citep{wu2023timesnet},  temporal 2D-variation modeling approach for general time series analysis.
    
\end{itemize}

\section{Verbalized Feature Extraction Plans}\label{sec:pan}
\subsection{RCW}
\subsubsection{Prompt}
Given the task: Play the role of a marine biology expert: is there a right whale call in the record?
Please tell me what features or patterns do you plan to extract from input data to correctly complete this task?

\subsubsection{GPT-4}
To determine if there are right whale calls in a recording while playing the role of a marine biology expert, several key features and patterns can be extracted from the audio data. Here’s a plan to approach this task:

Frequency Analysis: Right whale calls typically fall within specific frequency ranges, often between 50 Hz to 500 Hz. Analyzing the frequency content of the audio can help identify segments where these calls might occur.

Temporal Patterns: Right whale calls have distinctive durations and temporal patterns. For example, up-calls usually last about one second and rise in frequency. Identifying these patterns can help differentiate right whale calls from other sounds.

Energy and Power: Examining the energy or power of the audio signal over time can help to detect calls, as vocalizations tend to have higher energy levels compared to background noise.

Spectral Characteristics: Spectrogram analysis can be very useful. Right whale calls exhibit specific shapes on spectrograms, such as the curved shape of up-calls. Machine learning models can be trained to recognize these shapes.

Harmonic Patterns: Whale calls often consist of harmonic series where multiple harmonics of a fundamental frequency are present. Detecting these can help confirm the presence of whale calls.

Temporal Consistency: Repeated patterns over time could indicate the presence of biological sources (like whales) as opposed to random noise or other transient sounds.

Machine Learning Classification: After extracting features, employing a classifier such as a Support Vector Machine (SVM) or a neural network that has been trained on labeled examples of right whale calls and other sounds can automate the detection process.

Audio Quality Assessment: Ensuring the recording's quality and clarity can aid in more accurate analysis, considering factors like signal-to-noise ratio and possible interference.

By combining these methods, the task of detecting right whale calls in recordings can be approached systematically and effectively.
\subsubsection{GPT-4o}
To detect right whale calls in a sound recording, the following features and patterns are essential for accurate analysis. I will approach this task systematically, as it typically involves bioacoustic signal processing. Below are the key features and patterns I would focus on:

1. Spectral Features (Frequency Domain)
Spectrogram Analysis: Visual representation of frequencies over time to identify characteristic call patterns.
Peak Frequency: Right whale calls often occur between 50-300 Hz.
Band Energy Ratios: Measure energy distribution within specific frequency bands of interest.
Mel-Frequency Cepstral Coefficients (MFCCs): Captures timbral features, useful for bioacoustic identification.
2. Temporal Features (Time Domain)
Duration of Calls: Right whale calls typically last 0.5 - 1.5 seconds.
Temporal Gaps: Identifying rhythmic patterns or repetitions of the same call type.
Zero-Crossing Rate: Number of times the signal crosses the zero amplitude axis, useful for characterizing specific call types.
3. Call Types and Patterns
Upcalls: A rising frequency contour starting around 50 Hz and ending at 150-200 Hz (most common call type for North Atlantic right whales).
Gunshots: Brief, broadband pulses that are associated with social behavior or communication.
Chirps: Shorter calls with a rapid change in frequency.
Harmonics: Identify overtones accompanying the primary frequency of calls.
4. Acoustic Context Features
Ambient Noise: Identify and filter background noise like boat engines, wind, or waves to focus on whale vocalizations.
Signal-to-Noise Ratio (SNR): Assess how clearly the whale calls are distinguishable from background noise.
Call Repetition Patterns: Determine if calls appear in sequence or clusters, which is typical in some behavioral contexts.
5. Geospatial and Seasonal Context (if metadata is available)
Time of Day / Season: Right whales exhibit seasonal calling behavior—more frequent calls during migration and breeding seasons.
Location and Habitat Information: Helps determine if the recording region overlaps with known whale migration routes or habitats.
6. Feature Engineering and Post-processing Techniques
Smoothing and Filtering: Applying bandpass filters to isolate frequencies relevant to whale calls.
Dynamic Time Warping (DTW): Useful for matching whale call templates to observed audio signals.
Call Classification Models: Using supervised machine learning or deep learning models (e.g., Convolutional Neural Networks) trained on annotated whale call data.
By extracting and analyzing these features, I can determine the presence of right whale calls within a recording. The focus will be on identifying the upcall pattern, as it is the most diagnostic feature for right whales, but other patterns like gunshots and chirps will also be investigated to ensure comprehensive detection.
\subsubsection{Qwen2.5-72B}
Certainly! To determine whether a recording contains right whale calls, we need to analyze the acoustic characteristics of the audio data. Here’s a detailed breakdown of the features and patterns I would extract from the input data:

1. Frequency Analysis
Dominant Frequency Range: Right whales typically produce low-frequency vocalizations, often in the range of 50-200 Hz. Analyzing the frequency spectrum to identify dominant frequencies within this range can help confirm the presence of right whale calls.
Spectral Peaks: Look for distinct peaks in the frequency spectrum that correspond to known right whale call frequencies.
2. Temporal Analysis
Duration: Right whale calls can vary in duration but are generally between 0.5 and 2 seconds. Measuring the duration of potential calls can help filter out non-whale sounds.
Inter-Call Intervals: The intervals between successive calls can also be a useful feature. Right whales often produce calls at regular intervals, which can be characteristic of their vocal behavior.
3. Amplitude Modulation
Envelope Shape: Analyze the amplitude envelope of the sound to identify patterns typical of right whale calls. These calls often have a distinctive rise and fall in amplitude.
Peak Amplitude: Identify the peak amplitude of the sound and compare it to known thresholds for right whale calls.
4. Spectral Features
Spectral Centroid: This is the center of mass of the spectrum and can help distinguish between different types of sounds. Right whale calls tend to have a lower spectral centroid compared to other marine sounds.
Spectral Bandwidth: Measure the width of the frequency band where most of the energy is concentrated. Right whale calls often have a relatively narrow bandwidth.
5. Mel-Frequency Cepstral Coefficients (MFCCs)
MFCCs: These coefficients capture the shape of the spectral envelope and are widely used in speech and bioacoustic analysis. Extracting MFCCs from the audio can help in identifying the unique spectral patterns of right whale calls.
6. Waveform Analysis
Zero-Crossing Rate: The rate at which the waveform crosses the zero axis can provide information about the texture of the sound. Right whale calls often have a specific zero-crossing rate.
Energy Distribution: Analyze the energy distribution over time to identify segments with high energy that could correspond to whale calls.
7. Machine Learning and Pattern Recognition
Feature Vectors: Combine the above features into a feature vector for each segment of the audio.
Classification Models: Train machine learning models (e.g., Support Vector Machines, Neural Networks) on labeled data to classify segments as either containing right whale calls or not.
8. Contextual Analysis
Environmental Noise: Consider the ambient noise level and other marine sounds that might interfere with the detection of right whale calls.
Geographic and Temporal Context: Use metadata such as location and time of recording to cross-reference with known right whale habitats and migration patterns.
By extracting and analyzing these features, we can effectively determine whether a given audio recording contains right whale calls. This approach combines both traditional signal processing techniques and modern machine learning methods to ensure accurate identification.
\subsubsection{Qwen2-VL-72B}
To determine if there are right whale calls in a given audio record, I would extract the following features or patterns from the input data:

Spectrogram: A spectrogram is a visual representation of the spectrum of frequencies in a sound as they vary with time. By analyzing the spectrogram of the audio record, I can identify the frequency range and duration of the calls, which can help me distinguish right whale calls from other sounds.
Pitch: Right whale calls have a distinct pitch that is typically between 20 and 30 Hz. By analyzing the pitch of the audio record, I can determine if there are any sounds that fall within this range.
Duration: Right whale calls typically last between 1 and 10 seconds. By analyzing the duration of the sounds in the audio record, I can identify any sounds that fall within this range.
Intensity: Right whale calls are typically loud and can be heard over long distances. By analyzing the intensity of the sounds in the audio record, I can determine if there are any sounds that are loud enough to be right whale calls.
Repetition: Right whale calls are often repeated in a specific pattern. By analyzing the repetition of the sounds in the audio record, I can determine if there are any sounds that follow a pattern that is consistent with right whale calls.
By analyzing these features and patterns, I can determine if there are any right whale calls in the audio record.
\subsection{TEE}
\subsubsection{Prompt}
Given the task: Based on the power density time series data and select the transient electromagnetic event that best matches. The FORTE satellite detects transient electromagnetic events associated with lightning using a suite of optical and radio-frequency (RF) instruments. There are 7 event types. CG Positive Initial Return Stroke: A positive charge is lowered from a cloud to the ground. The characteristic feature of this type of event in the power density time series is a sharp turn-on of radiation, followed by a few hundreds of microseconds of noise; IR Negative Initial Return Stroke: A negative charge is lowered from a cloud to ground. The power waveform slowly ramps up to a level known as an attachment point, where a large surge current causes the VHF power to 'spike'. This attachment is followed by an exponentially shaped decline in the waveform.; SR Subsequent Negative Return Stroke: A negative charge is lowered from a cloud to ground. As the name implies, subsequent return strokes come after initial return strokes. Note that subsequent positive return strokes don't exist. I Impulsive Event: Typically an intra-cloud event characterized by a sudden peak in the waveform. I2 Impulsive Event Pair: Another intra-cloud event characterized by sudden peaks in the waveform that come in closely separated pairs. These are also called TIPPs (Trans-Ionospheric Pulse Pairs). KM Gradual Intra-Cloud Stroke: An intra-cloud event which increases in power more gradually than an impulsive event. O Off-record: 800 microseconds was not enough to fully capture the lightning event.
Please tell me what features or patterns do you plan to extract from input data to correctly complete this task?

\subsubsection{GPT-4}
To successfully classify transient electromagnetic events based on power density time series data from the FORTE satellite, you would analyze specific features from the time series that correspond to the unique characteristics of each event type. Here are the key features and patterns to focus on for each of the seven event types:

CG Positive Initial Return Stroke:

Sharp Onset of Radiation: Look for a very quick rise in power density at the beginning of the event.
Short Duration Noise: After the sharp onset, expect a period of noisy signal lasting a few hundred microseconds.
IR Negative Initial Return Stroke:

Slow Ramp-Up: The power density will increase gradually until it reaches a specific threshold.
Spike at Attachment Point: After the slow ramp-up, look for a significant spike in the power density.
Exponential Decline: Post-spike, the waveform should show an exponentially shaped decline.
SR Subsequent Negative Return Stroke:

Multiple Peaks: Since these strokes occur after the initial return strokes, identify multiple peaks that might follow initial peaks in a given timeframe.
Characteristics Similar to IR Negative: Each subsequent return stroke might mirror the slow ramp-up and sharp spike, though typically less pronounced than the initial stroke.
I Impulsive Event:

Sudden Peak: Look for a sudden, sharp peak in the power density without prior gradual increase or subsequent pairs of peaks.
I2 Impulsive Event Pair:

Paired Peaks: Identify closely spaced pairs of sharp peaks.
Consistency in Time Interval: The time interval between the paired peaks should be consistent across events classified as TIPPs.
KM Gradual Intra-Cloud Stroke:

Gradual Increase in Power: Unlike impulsive events, these will show a more gradual rise in power density.
Sustained High Power Levels: The power might stay elevated for a longer period compared to other intra-cloud events.
O Off-record:

Incomplete Waveform: Look for waveforms that seem to cut off or end abruptly without resolving normally within the 800-microsecond timeframe.
To automate the classification process and improve accuracy, you can implement a series of steps:

Preprocessing: Apply noise reduction and normalization techniques to clean the data for more precise analysis.
Feature Extraction: Develop algorithms to extract the above features from the time series data. This might include detecting peaks, analyzing the rate of rise and fall in power density, and measuring durations and intervals.
Classification Model: Use machine learning techniques such as decision trees, support vector machines, or neural networks to classify events based on the extracted features. Training the model with labeled examples of each event type will be crucial.
Validation and Testing: Continuously validate the model with new data and adjust parameters to handle variations in signal characteristics or noise levels.
By focusing on these features and employing robust data processing and machine learning techniques, you can effectively classify the types of transient electromagnetic events detected by the FORTE satellite.
\subsubsection{GPT-4o}
To correctly classify transient electromagnetic events detected by the FORTE satellite using the power density time series, I will extract the following features and patterns from the data:

1. Onset Shape and Slope (Turn-On Pattern)
Feature: Detect the rate of power increase at the start of the event.
Reason:
CG Positive Initial Return Stroke: Sharp, sudden increase in power.
IR Negative Initial Return Stroke: Gradual ramp-up to an attachment point, followed by a sudden spike.
2. Noise Characteristics (Post-Onset)
Feature: Identify high-frequency noise following the event onset.
Reason:
CG Positive Initial Return Stroke: Hundreds of microseconds of noise following the sharp turn-on.
IR Negative Initial Return Stroke: Less noise following the attachment point.
3. Peak Power Amplitude and Timing
Feature: Measure the peak power and the time to reach the peak.
Reason:
I Impulsive Event: Single, sharp peak.
I2 Impulsive Event Pair (TIPPs): Two closely spaced peaks of similar amplitude.
SR Subsequent Negative Return Stroke: Peak power smaller than the initial stroke, coming after an initial event.
4. Power Decay Pattern (Post-Peak)
Feature: Examine the post-peak decay shape (exponential or not).
Reason:
IR Negative Initial Return Stroke: Exponential decline following the surge.
KM Gradual Intra-Cloud Stroke: Power decays gradually over time without a sharp drop.
5. Duration of the Event
Feature: Calculate the total event duration (from onset to the end of the waveform).
Reason:
O Off-Record: The event continues beyond the 800-microsecond recording window.
Other Events: Fit within the 800-microsecond window.
6. Number of Peaks and Time Separation Between Peaks
Feature: Identify the number of distinct peaks and measure time intervals between them.
Reason:
I2 Impulsive Event Pair (TIPPs): Two closely spaced peaks.
I Impulsive Event: Single peak without subsequent pulses.
SR Subsequent Negative Return Stroke: A second peak occurring after the initial return stroke.
7. Power Level at Attachment Point (For IR Events)
Feature: Detect the power level at the attachment point for initial return strokes.
Reason:
IR Negative Initial Return Stroke: The power ramps up gradually and hits an attachment point before spiking.
8. Pattern in Power Increase (Gradual vs. Sudden)
Feature: Classify the pattern of power increase (impulsive or gradual).
Reason:
KM Gradual Intra-Cloud Stroke: Gradual increase in power compared to impulsive events.
I Impulsive Event: Sudden increase in power.
These features will help in differentiating between the seven event types by matching the time series patterns against the characteristic behaviors described.
\subsubsection{Qwen2.5-72B}
To accurately classify transient electromagnetic events based on the power density time series data from the FORTE satellite, several key features and patterns should be extracted and analyzed. Here’s a detailed breakdown of what to look for in each event type:

1. CG Positive Initial Return Stroke
Sharp Turn-On: Look for a sudden and significant increase in power density over a very short period (a few microseconds).
Noise Duration: After the sharp turn-on, there should be a period of noise lasting a few hundred microseconds.
2. IR Negative Initial Return Stroke
Ramp-Up: Identify a gradual increase in power density leading to an attachment point.
Spiking: At the attachment point, there should be a sharp spike in the power waveform.
Exponential Decline: Following the spike, the power should decline exponentially.
3. SR Subsequent Negative Return Stroke
Sudden Peak: Look for a sudden and significant increase in power density, similar to the initial return stroke but typically less intense.
Timing: Consider the timing relative to the initial return stroke, as subsequent strokes often occur shortly after the initial one.
4. I Impulsive Event
Sudden Peak: Identify a sudden and sharp peak in the power waveform.
Duration: The peak should be relatively short-lived, typically a few microseconds.
5. I2 Impulsive Event Pair (TIPPs)
Paired Peaks: Look for two closely spaced peaks in the power waveform.
Separation: The separation between the peaks should be very short, typically a few microseconds.
6. KM Gradual Intra-Cloud Stroke
Gradual Increase: Identify a gradual and sustained increase in power density over a longer period compared to impulsive events.
Smooth Transition: The transition from low to high power should be smooth without sharp spikes.
7. O Off-record
Duration Exceeding 800 microseconds: If the event lasts longer than 800 microseconds, it is classified as off-record.
Feature Extraction Techniques:
Time Domain Analysis:
Peak Detection: Use algorithms to detect peaks in the power density waveform.
Rise Time: Measure the time it takes for the power to rise from a baseline to a peak.
Fall Time: Measure the time it takes for the power to fall from a peak back to a baseline.
Noise Analysis: Analyze the noise characteristics following a sharp turn-on.
Frequency Domain Analysis:
Spectral Analysis: Perform Fourier Transform to analyze the frequency components of the waveform.
Bandwidth: Determine the bandwidth of the signal, which can help distinguish between different types of events.
Statistical Analysis:
Mean and Variance: Calculate the mean and variance of the power density over different segments of the waveform.
Skewness and Kurtosis: These higher-order statistics can provide insights into the shape of the waveform.
Machine Learning:
Feature Engineering: Combine the above features into a feature vector.
Classification Models: Train machine learning models (e.g., Random Forest, Support Vector Machines, Neural Networks) to classify the events based on the extracted features.
By carefully extracting and analyzing these features, you can effectively classify the transient electromagnetic events detected by the FORTE satellite.
\subsubsection{Qwen2-VL-72B}
To correctly complete this task, I plan to extract the following features or patterns from the input data:

Sharp turn-on of radiation: This feature is characteristic of a positive initial return stroke (CG Positive Initial Return Stroke). I will look for a sudden increase in power density at the beginning of the time series data.
Exponentially shaped decline: This feature is characteristic of a negative initial return stroke (IR Negative Initial Return Stroke). I will look for a slow increase in power density followed by a sudden spike and then a gradual decline in the waveform.
Subsequent negative return stroke: This feature is characteristic of a subsequent negative return stroke (SR Subsequent Negative Return Stroke). I will look for a negative charge being lowered from a cloud to the ground after an initial return stroke.
Sudden peak in the waveform: This feature is characteristic of an impulsive event (I Impulsive Event). I will look for a sudden increase in power density in the time series data.
Closely separated pairs of sudden peaks: This feature is characteristic of an impulsive event pair (I2 Impulsive Event Pair). I will look for two closely separated peaks in the time series data.
Gradual increase in power: This feature is characteristic of a gradual intra-cloud stroke (KM Gradual Intra-Cloud Stroke). I will look for a gradual increase in power density in the time series data.
Off-record event: This feature is characteristic of an off-record event (O Off-record). I will look for a lightning event that was not fully captured within the 800 microseconds of the time series data.
\subsection{ECG}
\subsubsection{Prompt}
Given the task: As a cardiologist, you are tasked with classifying a patient's heart condition based on single-lead ECG recordings.
Please tell me what features or patterns do you plan to extract from input data to correctly complete this task?

\subsubsection{GPT-4}
To classify a patient's heart condition based on single-lead ECG recordings effectively, various features and patterns can be extracted from the ECG signal to facilitate accurate diagnosis. Here are some key features and patterns typically considered:

Heart Rate: The average heart rate can be calculated by detecting the intervals between R-peaks (RR intervals) in the ECG signal. Variations in heart rate can indicate conditions like tachycardia or bradycardia.

RR Intervals: Analyzing the variability of RR intervals helps in assessing the autonomic nervous system's control over the heart, indicating potential arrhythmias or other heart conditions.

P-Wave Analysis: The presence, size, shape, and duration of the P-wave, which represents atrial depolarization, are important. Abnormalities in P-waves can indicate atrial enlargement or atrial arrhythmias.

QRS Complex: The duration, amplitude, and morphology of the QRS complex, which represents ventricular depolarization, are crucial. Changes can indicate ventricular hypertrophy, bundle branch blocks, or other ventricular disorders.

ST Segment: The level and shape of the ST segment can indicate ischemia or myocardial infarction. Elevation or depression of this segment is particularly significant in diagnosing these conditions.

T-Wave Analysis: Alterations in T-wave morphology can be indicative of electrolyte imbalances, ischemia, or myocardial infarction.

QT Interval: Measuring the duration of the QT interval, which represents the total time for ventricular depolarization and repolarization, is important. Prolonged or shortened QT intervals can lead to arrhythmias.

Signal Quality: Assessing the quality of the ECG signal to detect noise, artifacts, or missing segments which could affect the analysis.

Advanced Signal Processing Features:

Spectral Analysis: Frequency components of the ECG can provide insights into periodic oscillations of the heart rhythm, identifying arrhythmic conditions.
Wavelet Transform: This helps in detecting transient features and non-stationary changes in the ECG signal.
Machine Learning Features:

Feature Engineering: Creating composite features like heart rate variability, RR interval statistics (mean, median, range, standard deviation), and counts of arrhythmic beats.
Time-Series Analysis: Applying algorithms to detect trends, patterns, and outliers over time.
Statistical Features: These include calculating the mean, variance, skewness, and kurtosis of the intervals and amplitudes, providing a statistical summary that may indicate underlying pathologies.

These features can be extracted using various signal processing techniques and then used as inputs into classification models or algorithms to determine specific heart conditions. By analyzing these aspects of the ECG, a cardiologist can effectively classify different types of heart conditions with higher accuracy.
\subsubsection{GPT-4o}

To classify a patient’s heart condition based on single-lead ECG recordings, the following key features and patterns can be extracted:

1. Time-Domain Features
Heart Rate (HR):

Average beats per minute (bpm) over the recording.
Heart rate variability (HRV) to assess autonomic function.
R-R Interval:

Interval between successive R-peaks, essential for rhythm assessment.
P, QRS, and T Wave Duration:

Abnormal durations may indicate conduction delays (e.g., bundle branch block).
PR Interval:

Prolonged PR intervals could indicate first-degree heart block.
QT and Corrected QT Interval (QTc):

Long QT may indicate risk for arrhythmias.
QRS Complex Morphology:

Shape or width of the QRS complex provides insight into ventricular conduction.
Frequency-Domain Features
Power Spectral Density (PSD):

Analysis of frequency components to detect anomalies in heart rhythms (e.g., tachyarrhythmia).
LF/HF Ratio:

Low-Frequency (LF) and High-Frequency (HF) ratio reflects autonomic nervous system balance.
Morphological Features
Amplitude of P, QRS, and T Waves:

Useful to detect hypertrophy, ischemia, or electrolyte imbalances.
ST-Segment Deviation:

Elevated or depressed ST-segment indicates ischemia or myocardial infarction.
Presence of U-Waves:

Might indicate hypokalemia or repolarization abnormalities.
Rhythm Patterns and Waveforms
Detection of Arrhythmias:

Identification of atrial fibrillation (AF), ventricular tachycardia (VT), or bradycardia.
Atrial and Ventricular Premature Contractions:

Isolated ectopic beats that may suggest arrhythmogenic risks.
Regularity of Beats:

Distinguishing between sinus rhythm, AF, and other irregular rhythms.
Nonlinear and Statistical Features
Poincaré Plots:

Used to visualize HRV patterns and detect arrhythmias.
Entropy (e.g., Approximate and Sample Entropy):

Measures signal complexity, with reduced entropy indicating abnormal rhythms.
Fractal Dimension:

Provides insight into the complexity of ECG signals, useful for arrhythmia detection.
Wavelet Transform Features
Wavelet Coefficients:

Helps in time-frequency analysis to detect transient events like arrhythmias or ischemia.
Multi-scale Analysis:

Captures changes in the ECG signal across different scales, useful for feature extraction from noisy signals.
Machine Learning or Deep Learning Based Features (if applicable)
Learned Features from CNN or RNN Models:

Automatically extracted patterns that represent ECG morphology and rhythm.
Latent Features from Autoencoders:

Captures compressed representations of the signal for efficient classification.
By extracting these features, both traditional cardiologists and machine learning models can perform accurate heart condition classification based on ECG recordings, improving the diagnostic process and patient outcomes.
\subsubsection{Qwen2.5-72B}
To classify a patient's heart condition based on single-lead ECG recordings as a cardiologist, several key features and patterns need to be extracted from the ECG data. These features can help in diagnosing various cardiac conditions such as arrhythmias, ischemia, and myocardial infarction. Here are the primary features and patterns to consider:

Heart Rate (HR):
Mean Heart Rate: Average beats per minute (BPM).
Heart Rate Variability (HRV): Variations in the time interval between heartbeats.
QRS Complex:
Amplitude: The height of the QRS complex.
Duration: The width of the QRS complex.
Morphology: Shape and pattern of the QRS complex, which can indicate bundle branch blocks or ventricular hypertrophy.
P Wave:
Amplitude: Height of the P wave.
Duration: Width of the P wave.
Morphology: Shape and pattern of the P wave, which can indicate atrial enlargement or atrial fibrillation.
T Wave:
Amplitude: Height of the T wave.
Duration: Width of the T wave.
Morphology: Shape and pattern of the T wave, which can indicate ischemia or electrolyte imbalances.
ST Segment:
Elevation or Depression: Changes in the ST segment can indicate myocardial ischemia or infarction.
Morphology: Shape of the ST segment, including upsloping, downsloping, or flat patterns.
PR Interval:
Duration: Time from the onset of the P wave to the onset of the QRS complex.
Variability: Changes in the PR interval can indicate conduction abnormalities.
QT Interval:
Duration: Time from the onset of the Q wave to the end of the T wave.
Corrected QT (QTc): Adjusted for heart rate to account for variations in heart rate.
RR Interval:
Duration: Time between successive R waves.
Variability: Changes in RR intervals can indicate arrhythmias.
Rhythm Analysis:
Regular vs. Irregular: Consistency of the R-R intervals.
Rate: Whether the heart rate is normal, bradycardic, or tachycardic.
Pattern: Specific patterns such as atrial fibrillation, atrial flutter, or ventricular tachycardia.
Waveform Analysis:
Presence of Abnormal Waves: Such as U waves, epsilon waves, or delta waves.
Abnormal Patterns: Such as J-point elevation, Osborn waves, or T-wave inversions.
Frequency Domain Analysis:
Power Spectral Density (PSD): Distribution of power over different frequency bands.
Dominant Frequencies: Frequencies with the highest power.
Time-Frequency Analysis:
Spectral Analysis: Techniques like Fourier Transform or Wavelet Transform to analyze both time and frequency components.
Bispectral Analysis: Higher-order spectral analysis to detect nonlinear interactions.
Machine Learning Features:
Feature Engineering: Creating new features from raw ECG data, such as statistical measures, derivatives, and integrals.
Feature Selection: Using techniques like PCA (Principal Component Analysis) or LASSO (Least Absolute Shrinkage and Selection Operator) to select the most relevant features.
By extracting and analyzing these features, you can effectively classify a patient's heart condition and provide an accurate diagnosis. Each feature provides a piece of the puzzle, and together they form a comprehensive picture of the patient's cardiac health.
\subsubsection{Qwen2-VL-72B}
To classify a patient's heart condition based on single-lead ECG recordings, I would extract the following features or patterns from the input data:

Heart rate: The number of heartbeats per minute can provide important information about the patient's heart condition.
Rhythm: The regularity of the heartbeats can indicate whether the patient has a normal or abnormal heart rhythm.
P wave: The P wave represents atrial depolarization and can provide information about the patient's atrial activity.
QRS complex: The QRS complex represents ventricular depolarization and can provide information about the patient's ventricular activity.
ST segment: The ST segment represents the period between the end of ventricular depolarization and the beginning of ventricular repolarization and can provide information about the patient's myocardial ischemia.
T wave: The T wave represents ventricular repolarization and can provide information about the patient's ventricular repolarization.
QT interval: The QT interval represents the time between the start of ventricular depolarization and the end of ventricular repolarization and can provide information about the patient's ventricular repolarization.
PR interval: The PR interval represents the time between the start of atrial depolarization and the start of ventricular depolarization and can provide information about the patient's atrioventricular conduction.
QRS duration: The QRS duration represents the time it takes for the ventricles to depolarize and can provide information about the patient's ventricular conduction.
Axis: The axis represents the direction of the electrical activity of the heart and can provide information about the patient's heart position and orientation.
By analyzing these features or patterns, I can classify the patient's heart condition based on the single-lead ECG recordings.
\subsection{EMG}
\subsubsection{Prompt}
Given the task: As an Electromyograms (EMG) analysis expert, you are tasked with determining the type of the subject based on the EMG record.
Please tell me what features or patterns do you plan to extract from input data to correctly complete this task?
\subsubsection{GPT-4}
As an Electromyograms (EMG) analysis expert tasked with determining the type of subject based on the EMG record, the following features and patterns would be essential to extract from the input data to facilitate accurate classification:

Signal Amplitude: The peak amplitude of the EMG signal provides information on muscle activity level, which can vary significantly between different types of subjects, such as athletes vs. non-athletes, or among different medical conditions.

Mean Absolute Value (MAV): This feature represents the average of the absolute values of the EMG signal. It is useful for estimating the overall muscle activation over time.

Variance: The variance of the EMG signal can help in assessing the signal's power and muscle fatigue, which may differentiate between subject types based on their endurance and muscle condition.

Root Mean Square (RMS): This is a measure of the signal's power, reflecting muscle force and fatigue. It's particularly useful in continuous monitoring of muscle activity.

Zero Crossing Rate (ZCR): This measures the rate at which the signal changes from positive to negative and vice versa, indicating muscle fiber recruitment patterns and firing rates.

Waveform Length: The cumulative length of the waveform over time, reflecting the complexity of the muscle activation pattern. It can indicate the contractile characteristics of different muscle groups.

Frequency Domain Features:

Median Frequency (MDF): This frequency divides the spectrum into two regions with equal power; it shifts downwards as muscles fatigue.
Mean Frequency (MNF): This is the average frequency weighted by the amplitude, used to assess muscle fatigue and fiber composition.
Power Spectral Density (PSD): Analysis of the distribution of power across various frequency bands can indicate the type of muscle activity and its intensity.
Entropy: This measures the complexity or randomness of the EMG signal, useful for distinguishing between controlled and uncontrolled muscle activity.

Higher Order Statistics (HOS): Skewness and kurtosis of the EMG signal provide insights into the symmetry and peakiness of the distribution, which can vary with different types of muscle activation.

Autoregressive Model Coefficients: Parameters from fitting an autoregressive model to the EMG signal can help in characterizing the muscle activity and can be used as features for classification.

Signal Decomposition:

Wavelet Transform: Decomposing the signal into wavelets to capture both frequency and location information about muscle activity.
Empirical Mode Decomposition (EMD): This non-linear and non-stationary signal analysis technique can adaptively decompose an EMG signal into intrinsic mode functions (IMFs), revealing hidden patterns.
Pattern Recognition: Detecting specific patterns of muscle activation that are characteristic of certain actions or types of subjects, using sequence modeling or neural networks.

These features can be extracted using advanced signal processing techniques and subsequently used in machine learning models to classify different types of subjects based on their EMG records. The choice of features and model depends on the specificity of the subjects being classified and the quality and type of EMG data available.
\subsubsection{GPT-4o}
Time-Domain Features
Mean Absolute Value (MAV):

Average of the absolute EMG signal, reflecting muscle contraction intensity.
Root Mean Square (RMS):

Indicates the power of the signal and reflects muscle contraction strength.
Zero Crossing Rate (ZCR):

The number of times the signal changes sign, related to muscle activation and fatigue.
Slope Sign Changes (SSC):

Number of slope changes in the signal, which identifies muscle state changes.
Waveform Length (WL):

Cumulative length of the signal waveform, representing complexity and activity.
Integrated EMG (IEMG):

Summed absolute values of the EMG signal, used to measure muscle workload over time.
Frequency-Domain Features
Mean Frequency (MNF):

Average frequency content of the signal, which decreases during muscle fatigue.
Median Frequency (MDF):

Frequency at which the power spectrum is divided into two equal halves, commonly used for fatigue detection.
Power Spectral Density (PSD):

Quantifies the power across different frequency bands, helping to identify fatigue or abnormal muscle states.
Band Power (BP):

Signal energy in specific frequency bands (e.g., 10-500 Hz for EMG).
Spectral Moments:

Useful for characterizing the distribution of spectral content across the frequency spectrum.
Time-Frequency and Nonlinear Features
Short-Time Fourier Transform (STFT):

Analyzes how the frequency components of the signal change over time.
Wavelet Transform:

Provides multi-scale analysis to capture transient muscle activations and fatigue patterns.
Entropy Measures:

Sample Entropy, Approximate Entropy: Measures the signal complexity, with reduced entropy indicating fatigue or pathology.
Fractal Dimension:

Assesses signal irregularity, with higher values indicating more complex muscle activation patterns.
Muscle Activation Features
Onset and Offset Detection:

Time points when a muscle starts or stops contracting, useful for activity detection.
Co-activation Patterns:

Correlation between different muscle groups, helping in understanding coordination and motor control.
Muscle Synergy Analysis:

Identifying groups of muscles activated together to perform specific tasks.
Fatigue and Pathology Detection Features
Decrease in MNF and MDF Over Time:

Indicators of muscle fatigue as spectral content shifts to lower frequencies.
Increase in Signal Amplitude and Variability:

Pathological conditions may exhibit irregular or spastic muscle activity.
Motor Unit Action Potential (MUAP) Features:

MUAP size, shape, and firing rates provide insights into motor unit recruitment and muscle health.
Statistical Features
Skewness and Kurtosis:

Indicates asymmetry and peakedness of the signal distribution, respectively.
Variance and Standard Deviation:

Reflects the variability in muscle activity, useful for fatigue and abnormality detection.
Cross-Correlation:

Measures synchronization between signals from different muscles.
Machine Learning or Deep Learning Features (if applicable)
Learned Features from CNN, RNN, or LSTM Models:

Automatically extracted patterns representing complex muscle activity and fatigue states.
Latent Features from Autoencoders:

Compressed representations for efficient subject classification.
By extracting these features from the EMG signal, both traditional analysis and machine learning models can accurately classify the subject type, providing insights into muscle performance, fatigue, or pathology.
\subsubsection{Qwen2.5-72B}
To determine the type of the subject based on Electromyogram (EMG) records, several key features and patterns can be extracted from the EMG data. These features help in distinguishing between different types of muscle activities, which can be indicative of various conditions or states of the subject. Here are some of the most important features and patterns to consider:

Amplitude:
Mean Amplitude: The average amplitude of the EMG signal over a period of time.
Peak Amplitude: The maximum amplitude reached during the recording.
Frequency:
Power Spectrum Density (PSD): Analyzing the frequency content of the EMG signal can provide insights into the muscle's activation patterns.
Median Frequency (MDF): The frequency below which 50% of the total power is contained.
Mean Frequency (MNF): The average frequency of the power spectrum.
Time-Domain Features:
Root Mean Square (RMS): A measure of the magnitude of the EMG signal over a period of time.
Zero Crossings (ZC): The number of times the EMG signal crosses the zero line.
Waveform Length (WL): The cumulative length of the EMG signal over a period of time.
Slope Sign Changes (SSC): The number of times the slope of the EMG signal changes sign.
Frequency-Domain Features:
Fast Fourier Transform (FFT): Converts the time-domain signal into its frequency components.
Band Power: The power in specific frequency bands (e.g., low, medium, high).
Non-Linear Features:
Sample Entropy (SampEn): Measures the complexity and regularity of the EMG signal.
Approximate Entropy (ApEn): Another measure of signal complexity and regularity.
Motor Unit Action Potential (MUAP) Analysis:
MUAP Shape and Size: Characteristics of individual motor unit action potentials.
Firing Rate: The rate at which motor units are activated.
Spectral Moments:
First Spectral Moment (F1): The mean frequency.
Second Spectral Moment (F2): The variance of the frequency distribution.
Time-Frequency Analysis:
Short-Time Fourier Transform (STFT): Provides a time-varying spectral representation of the EMG signal.
Wavelet Transform: Analyzes the signal in both time and frequency domains, useful for non-stationary signals.
Statistical Features:
Skewness: Measures the asymmetry of the signal distribution.
Kurtosis: Measures the "tailedness" of the signal distribution.
Muscle Fatigue Indicators:
Fatigue Index (FI): The rate of change of the median frequency over time, indicating muscle fatigue.
Spectral Shift: Changes in the frequency content of the EMG signal as the muscle fatigues.
By extracting these features, you can build a comprehensive profile of the EMG signal, which can then be used to classify the type of the subject. For example, different muscle conditions (e.g., healthy vs. pathological), different levels of muscle activity (e.g., rest vs. contraction), or different types of movements (e.g., voluntary vs. involuntary) can be distinguished based on these features. Machine learning algorithms can be trained on these features to automate the classification process.
\subsubsection{Qwen2-VL-72B}
As an EMG analysis expert, I would extract the following features or patterns from the input data to determine the type of the subject:

Frequency content: The frequency content of the EMG signal can provide information about the type of muscle activity. For example, low-frequency components may indicate slow muscle contractions, while high-frequency components may indicate fast muscle contractions.
Amplitude: The amplitude of the EMG signal can provide information about the strength of the muscle contraction. Higher amplitudes may indicate stronger muscle contractions.
Duration: The duration of the EMG signal can provide information about the duration of the muscle contraction. Longer durations may indicate sustained muscle contractions.
Shape: The shape of the EMG signal can provide information about the type of muscle activity. For example, a burst of activity followed by a period of inactivity may indicate a twitch, while a sustained period of activity may indicate a tetanus.
Coherence: The coherence between different EMG signals can provide information about the coordination of muscle activity. High coherence may indicate coordinated muscle activity, while low coherence may indicate uncoordinated muscle activity.
Power spectral density: The power spectral density of the EMG signal can provide information about the distribution of power across different frequency bands. This can be used to identify specific muscle groups or types of muscle activity.
By analyzing these features or patterns, I can determine the type of the subject based on the EMG record.
\subsection{CTU}

\subsubsection{Prompt}
Given the task: Play as a computer energy consumption analysis expert, please correctly determine whether this computer is a desktop or a laptop based on the 24-hour power consumption data.
Please tell me what features or patterns do you plan to extract from input data to correctly complete this task?
\subsubsection{GPT-4}
To differentiate between a desktop and a laptop based on 24-hour power consumption data, you would focus on extracting and analyzing specific features or patterns that can indicate the type of device based on its energy usage profile. Here are several key features and patterns you might consider:

Total Daily Power Consumption:

Laptops typically consume less power than desktops due to their more energy-efficient components. Calculating the total power used over a 24-hour period could give an initial indication of the device type.
Power Consumption Patterns Over Time:

Analyze hourly or segment-wise power consumption. Laptops might show a more uniform consumption pattern, especially if they're left on but are in sleep or hibernate modes. Desktops might show a starker contrast between high consumption during active use and low consumption when turned off or in sleep mode.
Frequency and Duration of Power Spikes:

Desktops might exhibit higher power spikes during usage due to more powerful processors and peripherals compared to laptops. Observing how often and how long these spikes occur can be indicative.
Minimum Power Consumption Levels:

The minimum power levels (especially during inactive periods like nighttime) can be telling. Desktops might completely turn off (very low or zero consumption) or remain on higher consumption levels due to connected devices, whereas laptops typically have lower baseline consumption levels due to battery optimization features.
Presence of Battery Charging Patterns:

If data shows periodic drops and rises in power consumption that could correspond to a battery charging cycle, it’s likely a laptop. Desktops would not show this pattern unless a UPS or similar device is connected, which is less common.
Variability in Power Consumption:

Analyzing the variability and standard deviation in power usage over 24 hours can help distinguish between the two. Laptops generally have less variability in power consumption, while desktops might have greater fluctuations due to different modes of operation (idle, full power, sleep mode).
Response to Day and Night Cycles:

Depending on the usage patterns, if the device shows a significant reduction in power usage during typical sleeping hours, it might suggest a laptop which is often put into sleep mode automatically. Desktops might not show this pattern distinctly if they are left on for processes like downloads, updates, or backups during off-hours.
By analyzing these features and considering the context in which the device is used (e.g., home, office), you can infer with reasonable accuracy whether the device is a desktop or a laptop. Statistical and machine learning models can be applied to these features to automate the classification process, especially if you have a labeled dataset to train such models.
\subsubsection{GPT-4o}
To determine whether a computer is a desktop or a laptop based on its 24-hour power consumption data, I will extract the following features and patterns:

1. Base Power Consumption (Idle Usage)
Feature: Identify the minimum power level when the computer is idle.
Reason: Desktops typically have higher idle consumption due to larger components (e.g., GPUs, cooling systems). Laptops generally consume lower power at idle.
2. Power Consumption Peaks and Variability
Feature: Extract peak power usage and how frequently peaks occur throughout the day.
Reason: Desktops often have higher peak power consumption during intensive tasks (e.g., gaming, rendering). Laptops may exhibit more frequent but lower peaks, depending on workload.
3. Usage Patterns (On-Off Cycles)
Feature: Identify the duration and frequency of usage periods vs. idle or sleep periods.
Reason: Laptops tend to have more off/sleep cycles due to battery-saving features, while desktops often remain on for extended periods, especially if connected to UPS or used as servers.
4. Power Spikes During Start-Up (Boot Time)
Feature: Detect significant power increases indicating a system boot.
Reason: Laptops may exhibit small, brief spikes during start-up, while desktops might have more noticeable power jumps due to larger components.
5. Battery-Induced Cycles (Power Drops)
Feature: Look for repeating low power levels or abrupt power drops corresponding to battery-saving modes (if the system is unplugged).
Reason: Only laptops have battery discharge patterns, leading to identifiable low-power modes (like Sleep or Hibernate).
6. Cooling System Patterns (Fan Activity)
Feature: Identify fan activity or cooling-induced power surges.
Reason: Desktops typically have stronger fans or liquid cooling systems, leading to regular power surges when temperatures rise. Laptops have smaller fans, with less significant fluctuations.
7. Day-Night Power Trends (User Behavior)
Feature: Observe power usage during the night and early morning hours.
Reason: Desktops are sometimes left on overnight for background processes (e.g., downloads, backups), while laptops are more likely to enter sleep mode overnight.
8. Power Consumption Range
Feature: Compute the range of power consumption (Max - Min) throughout the day.
Reason: Desktops generally exhibit a broader power range (e.g., 50-200W or higher). Laptops have a narrower range (e.g., 5-50W).
These features and patterns will allow us to distinguish between desktop and laptop behaviors effectively. With proper statistical or machine learning models, these extracted features can be combined for a final classification.
\subsubsection{Qwen2.5-72B}
To determine whether a computer is a desktop or a laptop based on 24-hour power consumption data, I would focus on several key features and patterns that typically differentiate the two types of computers. Here’s how I would approach the analysis:

1. Average Power Consumption
Desktops generally have higher average power consumption compared to laptops because they have more powerful components and larger power supplies.
Laptops are designed for portability and efficiency, so their average power consumption is typically lower.
2. Peak Power Consumption
Desktops can have very high peak power consumption, especially during resource-intensive tasks like gaming, video rendering, or running multiple applications simultaneously.
Laptops have more limited peak power consumption due to their smaller form factor and battery constraints.
3. Power Consumption Variability
Desktops often show more variability in power consumption throughout the day, as they are more likely to be used for a wide range of tasks with varying power demands.
Laptops tend to have more consistent power consumption, especially when running on battery power, but can show variability when plugged in and performing intensive tasks.
4. Idle Power Consumption
Desktops usually have a higher idle power consumption because they have more components (e.g., multiple fans, larger power supplies) that consume power even when the system is not under load.
Laptops have lower idle power consumption due to their energy-efficient design and power-saving features.
5. Power Consumption Patterns During Sleep or Hibernate
Desktops may not enter deep sleep or hibernate states as frequently as laptops, leading to higher power consumption even when not in use.
Laptops are more likely to enter low-power states (sleep or hibernate) to conserve battery, resulting in very low power consumption during these periods.
6. Daily Usage Patterns
Desktops are often left on for extended periods, even overnight, leading to a more continuous power consumption pattern.
Laptops are more likely to be turned off or put into sleep mode when not in use, resulting in more frequent drops in power consumption.
7. Battery Cycles (if applicable)
If the data includes information about battery cycles, this can be a clear indicator:
Laptops will show battery charge and discharge cycles.
Desktops do not have batteries, so there will be no such cycles.
8. Time of Day Analysis
Desktops might show higher power consumption during typical working hours (9 AM to 5 PM) and lower consumption during nighttime.
Laptops might show more varied usage patterns, including evening and late-night usage, depending on the user's habits.
Example Analysis Steps:
Calculate Average Power Consumption: Compute the mean power consumption over the 24-hour period.
Identify Peak Power Consumption: Find the highest power consumption value and note the time it occurs.
Analyze Power Consumption Variability: Calculate the standard deviation of power consumption to understand how much it fluctuates.
Examine Idle Power Consumption: Look at the lowest power consumption values to estimate idle power.
Check for Low-Power States: Identify periods of very low power consumption (e.g., <10W) that might indicate sleep or hibernate states.
Review Daily Usage Patterns: Plot the power consumption over the 24-hour period to visualize usage patterns and identify any regular cycles.
By combining these features and patterns, you can make an informed decision about whether the computer is a desktop or a laptop.
\subsubsection{Qwen2-VL-72B}
To determine whether a computer is a desktop or a laptop based on 24-hour power consumption data, I would extract the following features or patterns from the input data:

Average power consumption: I would calculate the average power consumption of the computer over the 24-hour period. Desktop computers typically consume more power than laptops, so a higher average power consumption would suggest that the computer is a desktop.
Power consumption during peak hours: I would also look at the power consumption during peak hours, such as during the day when the computer is in use. Desktop computers tend to have higher power consumption during peak hours, while laptops may have lower power consumption during these times.
Power consumption during idle periods: I would also examine the power consumption during idle periods, such as overnight when the computer is not in use. Desktop computers may have higher power consumption during idle periods due to their larger size and more powerful components, while laptops may have lower power consumption during these times.
Power consumption spikes: I would look for any power consumption spikes in the data, which could indicate that the computer is performing a power-intensive task. Desktop computers may have more frequent and larger power consumption spikes due to their more powerful components, while laptops may have fewer and smaller spikes.
By analyzing these features and patterns in the power consumption data, I would be able to determine whether the computer is a desktop or a laptop.
\subsection{HAR}
\subsubsection{Prompt}
Given the task: As a human activity recognition expert, you are tasked with determining the type of activity performed by the subject based on the accelerometer record series along the x, y, and z axes over time .
Please tell me what features or patterns do you plan to extract from input data to correctly complete this task?

\subsubsection{GPT-4}
For the task of human activity recognition based on accelerometer data along the x, y, and z axes, the extraction of relevant features is crucial for accurately classifying the type of activity. Here are some typical features and patterns you might consider extracting from the accelerometer data:

Statistical Features:

Mean: Average value of acceleration for each axis.
Standard Deviation: Measure of the amount of variation or dispersion in the acceleration values.
Variance: Squared deviation of each point from the mean, indicating the spread of the acceleration data.
Median: The middle value of the data which divides the probability distribution into two equal halves.
Range: Difference between the maximum and minimum values in the acceleration data.
Interquartile Range (IQR): Measures the statistical dispersion as the difference between 25th and 75th percentiles.
Time-domain Features:

Root Mean Square (RMS): Indicates the magnitude of acceleration, computed as the square root of the average of the squares of the values.
Zero Crossing Rate (ZCR): Number of times the signal changes from positive to negative and vice versa, which can indicate the frequency of the activity.
Signal Magnitude Area (SMA): Integration of the magnitude of the acceleration over a window, giving a sense of the energy expenditure.
Time between Peaks: Interval time between local maxima in the acceleration data.
Frequency-domain Features:

Fast Fourier Transform (FFT): Transforming the data from time domain to frequency domain to analyze the frequency components.
Power Spectral Density (PSD): Indicates the power present in various frequency components.
Spectral Entropy: Measures the regularity and complexity of the frequency distribution.
Peak Frequency: The frequency with the maximum power, indicating the dominant frequency of movement.
Correlation-based Features:

Correlation between Axes: Measures how related the movements in different axes are, which can indicate coordinated motion patterns.
Geometrical Features:

Angles between Axes: Can help in understanding the orientation of the body in space.
Magnitude of Acceleration Vector: Calculated as the square root of the sum of the squares of x, y, and z components. It provides a holistic view of the acceleration independent of the direction.
Entropy-based Features:

Signal Entropy: Indicates the unpredictability or complexity of the acceleration signal.
By extracting these features from the accelerometer data, you can capture a comprehensive profile of the movements, which can then be fed into a machine learning model to classify different types of human activities such as walking, running, sitting, standing, etc.
\subsubsection{GPT-4o}
To determine the type of activity performed by a subject using accelerometer data, I would focus on extracting both statistical features and domain-specific patterns that capture relevant aspects of the movements. Below is a structured approach to the features I plan to extract:

1. Time-Domain Features
These features capture statistical properties directly from the raw time series data along the x, y, and z axes:

Mean: Average value of the accelerometer data for each axis.
Standard Deviation (STD): Measures variability in the signal.
Variance: Quantifies how much the data varies.
Min, Max, and Range: Captures the spread of the data values.
Skewness: Measures the asymmetry of the data distribution.
Kurtosis: Measures the "tailedness" of the data distribution.
Signal Magnitude Area (SMA): 

Energy: Sum of squared values for each axis divided by the series length.
Root Mean Square (RMS): Useful to measure overall signal intensity.
2. Frequency-Domain Features
Using Fourier Transform (or Wavelet Transform) to analyze the signal in the frequency domain helps identify periodic activities (e.g., walking, running).

Spectral Entropy: Measures randomness in the frequency spectrum.
Dominant Frequency: The frequency with the highest magnitude.
Power Spectral Density (PSD): Helps measure power in various frequency bands.
Peak Frequencies and Their Amplitudes: To identify key frequencies related to the activity.
3. Time-Frequency Features
Using short-time Fourier transform (STFT) or wavelet decomposition to extract:

Wavelet coefficients: Capture both time-localized frequency content.
Instantaneous frequency: Measures how frequency content changes over time.
4. Magnitude and Orientation Features
Some activities are better identified through the magnitude or orientation of movement:

Acceleration Magnitude: 
Magnitude

Mean and Variance of Magnitude: Stability of the signal magnitude over time.
Tilt Angles: Angles between the gravity vector and each axis.
Roll and Pitch Angles: Describe the subject's orientation.
5. Temporal and Statistical Patterns
These features capture the dynamics of the signal over time.

Zero Crossing Rate: Counts the number of times the signal crosses zero, indicating changes in direction.
Autocorrelation: Measures similarity of the signal with a delayed version of itself.
Number of Peaks: Useful for repetitive activities like walking or jogging.
Signal Peaks and Valleys: Identify critical motion events.
6. Segment-Level Features
Segmenting the data into windows (e.g., sliding windows of 1-3 seconds) ensures more granular feature extraction.

Features within sliding windows: Mean, STD, etc., within each window.
Window Overlap: To smooth transitions between activities.
7. Cross-Axis Features
These features capture interactions between the x, y, and z axes:

Correlation between axes: Helps detect synchronized movements along different directions.
Covariance between axes: Measures joint variability of axes.
8. Activity-Specific Patterns
Gait Cycle Detection: For walking and running patterns.
Step Detection: Counting steps to differentiate walking from running.
Stationarity Detection: Helps identify stationary periods, such as sitting or standing.
9. Feature Engineering with Domain Knowledge
Tilt thresholding: To detect specific postures like sitting, lying down, or standing.
Jerk computation: Time derivative of acceleration to detect sudden changes (e.g., jumping).
This diverse set of features—statistical, frequency-based, temporal, and activity-specific—helps build a robust model capable of accurately classifying activities such as walking, running, sitting, standing, and more complex activities.
\subsubsection{Qwen2.5-72B}
To accurately determine the type of activity performed by a subject based on accelerometer data, it's crucial to extract meaningful features that can capture the essence of the movements. Here are some key features and patterns you might consider extracting from the accelerometer record series along the x, y, and z axes:

1. Statistical Features
Mean: The average value of the acceleration in each axis.
Standard Deviation: Measures the variability or spread of the acceleration values.
Variance: Another measure of the spread of the data.
Maximum and Minimum Values: The highest and lowest values of acceleration in each axis.
Range: The difference between the maximum and minimum values.
Median: The middle value of the sorted acceleration data.
Interquartile Range (IQR): The range between the first quartile (25th percentile) and the third quartile (75th percentile).
2. Frequency Domain Features
Fast Fourier Transform (FFT): Converts the time-domain signal into the frequency domain to identify dominant frequencies.
Spectral Energy: The total energy in the frequency domain.
Spectral Entropy: Measures the disorder or randomness in the frequency domain.
Spectral Centroid: The center of mass of the spectrum.
Spectral Roll-off: The frequency below which a certain percentage (e.g., 85%) of the total spectral energy is contained.
3. Time-Domain Features
Zero Crossing Rate: The rate at which the signal crosses zero.
Mean Absolute Value (MAV): The average of the absolute values of the acceleration.
Root Mean Square (RMS): The square root of the mean of the squared values of the acceleration.
Signal Magnitude Area (SMA): The sum of the absolute values of the acceleration in all three axes.
Jerk: The rate of change of acceleration, calculated as the difference between consecutive acceleration values.
4. Correlation and Covariance
Cross-Correlation: Measures the similarity between the acceleration signals in different axes.
Covariance: Measures how much two signals vary together.
5. Temporal Patterns
Autocorrelation: Measures the correlation of the signal with a delayed copy of itself.
Lag: The time delay at which the autocorrelation is maximized.
Peak Detection: Identifies the peaks in the acceleration signal, which can be indicative of specific activities.
6. Energy Features
Total Energy: The sum of the squared values of the acceleration.
Band Energy: The energy in specific frequency bands.
7. Complexity and Entropy
Sample Entropy: Measures the complexity of the signal.
Approximate Entropy: Another measure of signal complexity.
8. Shape Features
Skewness: Measures the asymmetry of the distribution of the acceleration values.
Kurtosis: Measures the "tailedness" of the distribution.
9. Activity-Specific Features
Step Count: For walking or running, the number of steps can be a useful feature.
Stride Length: The distance covered in one step.
Cadence: The number of steps per minute.
10. Machine Learning Features
Principal Component Analysis (PCA): Reduces the dimensionality of the data while retaining most of the variance.
Feature Selection: Techniques like mutual information, chi-squared test, or recursive feature elimination can help select the most relevant features.
Example Feature Extraction Process:
Preprocessing:
Filtering: Apply low-pass or band-pass filters to remove noise.
Segmentation: Divide the continuous data into fixed-length windows (e.g., 2 seconds).
Feature Calculation:
For each window, compute the statistical, frequency, time-domain, and other features mentioned above.
Feature Normalization:
Normalize the features to ensure they are on a similar scale, which can improve the performance of machine learning models.
Model Training:
Use the extracted features to train a machine learning model (e.g., Random Forest, SVM, Neural Networks) to classify the activities.
By carefully selecting and combining these features, you can build a robust model for human activity recognition using accelerometer data.
\subsubsection{Qwen2-VL-72B}
To determine the type of activity performed by the subject based on the accelerometer record series along the x, y, and z axes over time, I would extract the following features or patterns from the input data:

Mean and standard deviation of the acceleration values along each axis: These features can help identify the overall movement and intensity of the activity.
Frequency domain features: I would perform a Fourier transform on the acceleration data to extract frequency domain features such as the dominant frequency, frequency bandwidth, and spectral entropy. These features can help identify the periodicity and complexity of the activity.
Time domain features: I would extract time domain features such as the root mean square (RMS), zero crossing rate, and autocorrelation. These features can help identify the smoothness and regularity of the activity.
Activity-specific features: Depending on the type of activity being performed, I would extract activity-specific features such as the number of steps taken, the duration of the activity, and the intensity of the activity.
Contextual features: I would also consider contextual features such as the location of the subject, the time of day, and the duration of the activity. These features can help identify the context in which the activity is being performed.
By combining these features, I can create a comprehensive representation of the activity being performed and use machine learning algorithms to classify the activity accurately.

\begin{table*}[]
\centering

\begin{tabular}{ccc}
\hline
    & Cost for Traditional Method Per Sample & Cost for Proposed Method Per Sample \\ \hline
RCW & 0.15\$                                  & 0.000657\$                          \\
TEE & 0.0112\$                               & 0.000213\$                          \\
ECG & 0.0638\$                               & 0.000213\$                          \\
EMG & 0.0638\$                               & 0.000213\$                          \\
CTU & 0.027\$                                & 0.000213\$                          \\
HAR & 0.0185\$                               & 0.000213\$                          \\ \hline
\end{tabular}
\caption{Results of Cost Comparison based on GPT-4o. RCW leverages frequency-domain visualizations and requires an automatic resolution option. The remaining datasets use the same plot size and resolution, so they have consistent token usage.}\label{exp_cost1}
\end{table*}

\begin{table*}
\centering
\begin{tabular}{ccc} 
\hline
    & Tokens Per Sample for Numerical Modeling & Tokens Per Sample for Proposed Solution  \\ 
\hline
RCW & 60,000                                        & 262
                                                      \\
TEE & 4,466                                         & 85                                                      \\
ECG & 25,500                                        & 85                                                      \\
EMG & 25,500                                        & 85                                                      \\
CTU & 10,800                                        & 85                                                      \\
HAR & 7,416                                         & 85                                                      \\
\hline
\end{tabular}
\caption{Details of Token Usage Comparison based on GPT-4o. The remaining datasets use the same plot size and resolution, so they have consistent token usage.}\label{exp_cost2}
\end{table*}
 \begin{figure}[!ht]
     \centering
     \includegraphics[width=\linewidth]{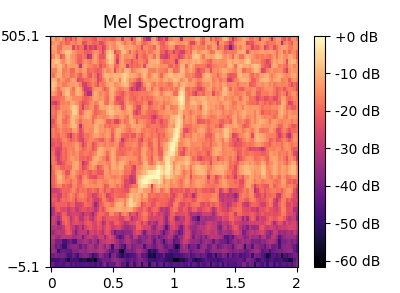}
     \caption{Frequency-Domain Visualization Demo of RCW Task.}
     \label{fig:RCW1}
\end{figure}
 \begin{figure}[!ht]
     \centering
     \includegraphics[width=\linewidth]{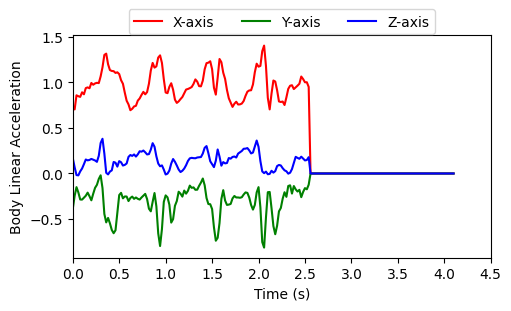}
     \caption{Time-Domain Visualization Demo of HAR Task.}
     \label{fig:HAR}
\end{figure}